\documentclass{article} 
\usepackage{iclr2025_conference,times}

\usepackage{amsmath,amsfonts,bm}

\def\eqref#1{equation~\ref{#1}}

\def\1{\bm{1}}

\DeclareMathAlphabet{\mathsfit}{\encodingdefault}{\sfdefault}{m}{sl}
\SetMathAlphabet{\mathsfit}{bold}{\encodingdefault}{\sfdefault}{bx}{n}

\usepackage{hyperref}
\usepackage{url}
\usepackage{xcolor}
\definecolor{darkpink}{RGB}{255, 20, 147}
\usepackage{makecell}
\usepackage{tabularx}
\usepackage{algorithm}
\usepackage{colortbl}
\usepackage{multirow}
\usepackage{multicol}
\usepackage{url}
\usepackage{adjustbox}
\usepackage{bbding} 
\usepackage{graphicx}
\usepackage{subcaption}
\usepackage{amsmath}
\usepackage{listings}
\usepackage{wrapfig}
\usepackage{float}
\usepackage{booktabs}
\usepackage{listings}
\title{Kronecker Mask and Interpretive Prompts are Language-Action Video Learners}

\author{Jingyi Yang\textsuperscript{\rm 1}$^*$, \quad Zitong Yu\textsuperscript{\rm 2\rm 3}\thanks{Equal Contribution}, \quad Xiuming Ni\textsuperscript{\rm 4}, \quad Jia He\textsuperscript{\rm 4}, \quad Hui Li\textsuperscript{\rm 1}\thanks{Corresponding Author} \\
\textsuperscript{\rm 1}University of Science and Technology of China, \quad \textsuperscript{\rm 2}Great Bay University, 
\\
\textsuperscript{\rm 3}Dongguan Key Laboratory for Intelligence and Information Technology, \\
\textsuperscript{\rm 4}Anhui Tsinglink Information Technology Co.,Ltd.\\
\texttt{yangjingyi@mail.ustc.edu.cn}, \quad \texttt{yuzitong@gbu.edu.cn},\\ \texttt{\{nixm,hejia\}@tsinglink.com}, \quad \texttt{mythlee@.ustc.edu.cn}\\
}

\iclrfinalcopy 
\begin{document}

\maketitle

\begin{abstract}
Contrastive language-image pretraining (CLIP) has significantly advanced image-based vision learning. A pressing topic subsequently arises: how can we effectively adapt CLIP to the video domain? Recent studies have focused on adjusting either the textual or visual branch of CLIP for action recognition. However, we argue that adaptations of both branches are crucial. In this paper, we propose \textbf{CLAVER}: a \textbf{C}ontrastive \textbf{L}anguage-\textbf{A}ction \textbf{V}ideo Learn\textbf{er}, designed to shift CLIP's focus from the alignment of static visual objects and concrete nouns to the alignment of dynamic action behaviors and abstract verbs. Specifically, we introduce a novel Kronecker mask attention for temporal modeling. Our tailored Kronecker mask offers three benefits 1) it expands the temporal receptive field for each token, 2) it serves as an effective spatiotemporal heterogeneity inductive bias, mitigating the issue of spatiotemporal homogenization, and 3) it can be seamlessly plugged into transformer-based models. Regarding the textual branch, we leverage large language models to generate diverse, sentence-level and semantically rich interpretive prompts of actions, which shift the model's focus towards the verb comprehension. Extensive experiments on various benchmarks and learning scenarios demonstrate the superiority and generality of our approach. Code is available at \href{https://github.com/yjyddq/CLAVER}{\textcolor{darkpink}{https://github.com/yjyddq/CLAVER}}.
\end{abstract}

\section{Introduction}
\label{sec:intro}
Video action recognition has long been a representative topic in video understanding. Over the past decade, there has been a continuous pursuit of learning spatiotemporal representations, giving rise to diverse architectures, such as traditional two-stream networks \cite{simonyan2014two,wang2016temporal,zhou2018temporal,karpathy2014large,xie2024fusionmamba}, 3D convolutional neural networks \cite{carreira2017quo,feichtenhofer2020x3d,feichtenhofer2019slowfast,hara2017learning,qiu2017learning,tran2015learning,tran2018closer,wang2018non,xie2018rethinking}, and Video Vision Transformers \cite{arnab2021vivit,bertasius2021space,fan2021multiscale,liu2022video,patrick2021keeping,zhao2022alignment,li2022uniformer,yan2022multiview}. Recently, there has been increasing interest in leveraging visual-language models (VLMs) like CLIP \cite{radford2021learning}, Florence \cite{yuan2021florence}, and ALIGN \cite{jia2021scaling} for various video tasks, owing to the superior generalization abilities of these models. Several studies \cite{wang2021actionclip,lin2022frozen,ni2022expanding,ju2022prompting,rasheed2023fine,tu2023implicit,chen2023video} have devoted to adapt the CLIP for video action recognition, but they often focus on adjusting a single branch. According to predecessor studies, transferring CLIP from the image domain to the video domain involves two key considerations: 1) how to perform effective temporal modeling. 2) how to design suitable text descriptions for verb understanding that align with rich text semantics in the VLM's pre-training dataset. We argue that addressing both issues simultaneously is crucial.

In addressing the issue 1), several studies \cite{wang2021actionclip,ju2022prompting,chen2023video,rasheed2023fine} implement straightforward and simple strategies such as mean pooling or 1D-temporal convolution across the temporal dimension, or employing temporal attention among class tokens. X-CLIP \cite{ni2022expanding}  and CLIP-ViP \cite{xue2022clip} introduce extra tokens for cross-frame communication. Alternatively, some studies \cite{lin2022frozen,tu2023implicit} engineer tailored modules. In our work, we aim to elucidate the distinctions and intrinsic correlations between space and time, as well as design more general spatiotemporal modeling approaches. 


Regarding the issue 2), some studies \cite{hendricks2021probing,thrush2022winoground} indicate that VLMs tend to focus on the correspondence between visual objects and nouns rather than action behaviors and verbs. We consider the essential gap is that visual objects are static and presented in lower dimensions, and nouns are concrete and easily understandable. However, action behaviors are dynamic that presented in higher dimensions, and verbs are intricate and abstract. Several existing methods \cite{ni2022expanding,lin2022frozen,rasheed2023fine,tu2023implicit} use verbs or phrases as direct text descriptions. ActionCLIP \cite{wang2021actionclip} integrates prompt templates to expand verbs or phrases into sentences. Ju et al. \cite{ju2022prompting} propose trainable continuous prompts to construct virtual prompt templates. However, these methods do not address aforementioned issues in essence. Alternatively VFC \cite{momeni2023verbs} and MAXI \cite{lin2023match} consider leveraging large language models (LLMs) to provide positive and negative text samples for contrastive learning or multiple instance learning, while ASU \cite{chen2023video} presents the concept of semantic units to supplement the semantic information of action labels.

\begin{figure*}[t]
    \centering
    \includegraphics[width=0.98\textwidth,height=0.28\textwidth]{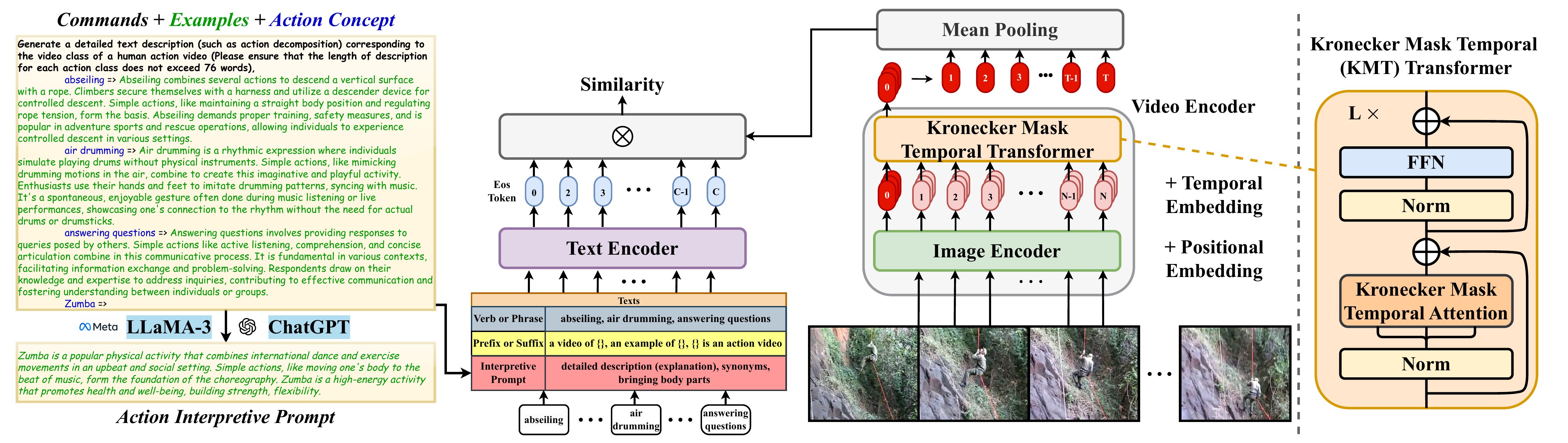}
    \caption{An overview of \textbf{CLAVER}. (\textbf{Right}) Image encoder and KMT transformer are assembled as a video encoder. (\textbf{Left}) How to get the interpretive prompts for actions.}
    \label{fig: Framework}
\vspace{-0.5cm}
\end{figure*}

To address these issues, we propose a \textbf{C}ontrastive \textbf{L}anguage-\textbf{A}ction \textbf{V}ideo Learn\textbf{er} (\textbf{CLAVER}, Fig. \ref{fig: Framework}) to efficiently adapt the CLIP for video action recognition. Specifically, for the issue 1), we first obtain the frame-level visual representation from the image encoder, then apply tailored Kronecker mask for temporal modeling with a wider temporal receptive field to establish long-range and wide-range dependencies among frames, while mitigating spatiotemporal homogenization. Additionally, we reveal the intrinsic correlations between space and time from the perspective of Kronecker mask attention. Regarding the issue 2), we leverage LLMs to effectively generate diverse, sentence-level, and semantically rich interpretations of actions, augmenting text descriptions during training and testing. This approach allows the text descriptions to be presented in a more flexible, sentence-level form during inference. In summary, our main contributions are four-fold:
\begin{itemize}
\item We propose the \textbf{C}ontrastive \textbf{L}anguage-\textbf{A}ction \textbf{V}ideo Learn\textbf{er} (\textbf{CLAVER}) to adapt both the visual and textual branches, efficiently shifting the alignment in CLIP from visual objects and nouns to action behaviors and verbs.
\item We propose the Kronecker mask temporal attention and Kronecker mask causal temporal attention for temporal modeling, aiming to capture the long-range and wide-range dependencies among frames with spatiotemporal heterogeneity. 
\item We introduce interpretive prompts of actions to facilitate the alignment of action behaviors and verbs, thereby improving zero-shot and few-shot generalization capabilities.
\item Extensive qualitative and quantitative experiments demonstrate the effectiveness of CLAVER. Our method achieves superior or competitive performance on Kinetics-400 and Kinetics-600 under fully-supervised scenario, and on HMDB-51 and UCF-101 under zero-shot, few-shot scenarios.
\end{itemize}

\section{Related Work}
\label{sec:related work}
\textbf{Video Recognition.} Among early video recognition methods, 3D convolution is widely employed \cite{qiu2017learning,tran2015learning,tran2018closer,xie2018rethinking,feichtenhofer2019slowfast,feichtenhofer2020x3d,yu2021searching}. Some studies \cite{qiu2017learning,tran2018closer,xie2018rethinking} propose to factorize convolutional operations across spatial and temporal dimensions, while others design the specific temporal modules to embed them into 2D CNNs \cite{li2020tea,lin2019tsm,liu2021tam}. Over the past few years, there has been an influx of transformer-based video works \cite{arnab2021vivit,neimark2021video,bertasius2021space,fan2021multiscale,liu2022video,yan2022multiview,li2022uniformer}, demonstrating promising performance. For example, some methods \cite{arnab2021vivit,neimark2021video,girdhar2021anticipative} adopt a factorized encoder structure for spatial-temporal fusion. Alternatively, another family employ a factorized attention structure. Such as TimeSformer \cite{bertasius2021space}, ViViT \cite{arnab2021vivit}, and ATA \cite{zhao2022alignment} which proposes alignment-guided tepmoral attention. While Video Swin Transformer \cite{liu2022video} employs 3D window attention. The representative attention calculation forms in these works can be summarized into joint attention and factorized (divided) spatiotemporal attention. In this paper, to make minimal modifications to the original structure, we adopt a factorized encoder structure for video stream. 

\noindent
\textbf{Visual-language Representation Learning.} Visual-textual multi-modality is a hot topic in recent years. Several studies based on masked image modeling (MIM) have achieved commendable performance \cite{li2020unicoder,lu2019vilbert,su2019vl,tan2019lxmert}. There are also efforts focused on video-language representation learning \cite{miech2019howto100m,sun2019learning,sun2019videobert,zhu2020actbert,xu2021videoclip}. Concurrently, contrastive language-image pretraining \cite{radford2021learning,jia2021scaling,yuan2021florence} achieved remarkable progress, particularly in demonstrating impressive zero-shot generalization capacities. CLIP \cite{radford2021learning} is one of the most representative works, with numerous follow-up studies have explored to adapt it for downstream tasks. For example, object detection, semantic segmentation, video retrieval and captioning, etc.\cite{gu2021open,vinker2022clipasso,li2022grounded,luo2022clip4clip,xu2022groupvit}. Additionally, there are also many applications in the video action recognition \cite{wang2021actionclip,ni2022expanding,lin2022frozen,ju2022prompting,pan2022st,rasheed2023fine,chen2023video,tu2023implicit,lin2023match,momeni2023verbs,yang2023aim}. For instance, ViFiCLIP \cite{rasheed2023fine} aim to minimize modifications to original models and facilitate efficient transfer, while Chen Ju et al. \cite{ju2022prompting} suggest optimizing a few prompt vectors for adapting CLIP to various video understanding tasks. X-CLIP \cite{ni2022expanding} proposes an efficient cross-frame attention module. ILA \cite{tu2023implicit} designs implicit mask-based alignment to align features of two adjacent frames and EVL \cite{lin2022frozen} proposes a image encoder and video decoder structure. Regrading to the language branch, most previous works directly use verbs or phrases that lack rich semantics which overlook the importance of semantics. With the advancement of large language models like the GPT-3 \cite{brown2020language}, PaLM \cite{chowdhery2023palm}, LLaMAs \cite{touvron2023llama,touvron2023llama2,dubey2024llama3herdmodels} and ChatGPT \cite{openai2024gpt4technicalreport}. LLMs can replace manual labor \cite{chen2021elaborative,qian2022rethinking} and automatically generate texts that meet human expectations to benefit visual-textual learning. For example, LaCLIP \cite{fan2024improving} employs LLMs to rewrite text descriptions associated with each image for text augmentation. VFC \cite{momeni2023verbs} and MAXI \cite{lin2023match} leverage LLMs to generate positive and negative texts with diversity for language-video learning. 


\section{Methodology}
\label{sec:methodology}
In Sec. \ref{sec:overview}, we overview our proposed contrastive language-action video learner architecture. Then, we elaborate on the detail of the Kronecker mask attention in Sec. \ref{sec:kronecker mask attention}. Finally, we present the technique details of action interpretive prompt in Sec. \ref{sec: interpretive prompt}.

\subsection{Overview}
\label{sec:overview}
Our contrastive language-action video learner architecture is illustrated in Fig. \ref{fig: Framework}. We utilize a video encoder to obtain video representations, comprising two transformers-based components: an image encoder (ViT) from CLIP, a Kronecker mask temporal transformer. The text encoder aims to align text representations with the video representations. Concretely, given a video clip \(\textit{V}=[v_0, \cdots v_t, \cdots v_{T-1}] \in \mathbb{R}^{T \times H \times W \times 3}, v_t \in \mathbb{R}^{H \times W \times 3}\) and corresponding text descriptions \(C=[c_0, \cdots c_m, \cdots c_{M-1}] \in \mathbb{R}^{M \times N}, c_m \in \mathbb{R}^{N}\), where \(T,H,W\) are the number of frames, height, width, respectively, \(M\) is the number of diverse text descriptions (share the same central concept) for an action category, \(N\) is the max sequence length. We feed texts \(C\) into the text encoder \(f_{\theta_C}(\cdot)\) to obtain text representations \(\textbf{C} = [\textbf{c}_0, \cdots \textbf{c}_{M-1}]\). For the video stream, firstly, we input the video clip \(V\) to the image encoder \(f_{\theta_I}(\cdot)\) to obtain frame-level representations \(\textbf{I}_t\).
\begin{equation}
\label{eqn: image encoder}
\textbf{I}_t = f_{\theta_I}(\mathrm{PE}(v_t)+\textbf{e}^{pos}), \quad \textbf{C} = f_{\theta_C}(C),
\end{equation}
where \(\mathrm{PE}(\cdot)\) is the patch embedding. Each frame is split into \(L=\frac{H}{P} \times \frac{W}{P}\) patches, \(L\) is the number of patches, \(P\) is the patch size, \(\textbf{e}^{pos}\) is the absolute positional embedding, \(\mathrm{PE}(v_t)+\textbf{e}^{pos} = [v_{t,0}+\textbf{e}^{pos}_0, \mathrm{PE}(v_{t,1})+\textbf{e}^{pos}_1, \cdots, \mathrm{PE}(v_{t,l})+\textbf{e}^{pos}_l, \cdots, \mathrm{PE}(v_{t,L})+\textbf{e}^{pos}_L] = [\textbf{z}_{t,0}, \cdots, \textbf{z}_{t,l}, \cdots, \textbf{z}_{t,L}]\),  \(v_{t,0}\) is the class token. \(\textbf{I}_t = f_{\theta_I}([\textbf{z}_{t,l}]_{t \in T, l \in L+1})= [\textbf{I}_{t,0}, \cdots, \textbf{I}_{t,l}, \cdots, \textbf{I}_{t,L}]\).

Then, we add absolute temporal embedding \(\textbf{e}^{tem}\) to the \(\textbf{I}_t,t \in T\) and feed them into the Kronecker mask temporal transformer \(f_{\theta_{V}}(\cdot)\). Finally, by selecting the class token from each frame and averaging them, we obtain a video representation \textbf{v} with the same dimension as \(\textbf{c}_m,m \in M\).
\begin{equation}
\label{eqn: video encoder}
\textbf{V} = f_{\theta_{V}}([\textbf{I}_t]_{t \in T}+\textbf{e}^{tem}), \quad \textbf{v} = \mathrm{Avg}([\textbf{V}_{t,0}]_{t \in T}),
\end{equation}
where \(\textbf{V} = f_{\theta_{V}}([\textbf{I}_t]_{t \in T} +\textbf{e}^{tem}) = f_{\theta_{V}}([\textbf{I}_t+\textbf{e}^{tem}_t]_{t \in T}) = 
f_{\theta_{V}}([[\textbf{I}_{0,0}+\textbf{e}^{tem}_0, \cdots, \textbf{I}_{0,L}+\textbf{e}^{tem}_0], \cdots, [\textbf{I}_{t,0}+\textbf{e}^{tem}_t, \cdots, \textbf{I}_{t,L}+\textbf{e}^{tem}_t], \cdots, [\textbf{I}_{T-1,0}+\textbf{e}^{tem}_{T-1}, \cdots, \textbf{I}_{T-1,L}+\textbf{e}^{tem}_{T-1}]]) =
[[\textbf{V}_{0,0}, \cdots, \textbf{V}_{0,L}], \cdots, [\textbf{V}_{t,0}, \cdots, \textbf{V}_{t,L}], \cdots, [\textbf{V}_{T-1,0}, \cdots, \textbf{V}_{T-1,L}]]\), \(\mathrm{Avg}(\cdot)\) is the average pooling function.
Our optimization goal is to maximize the cosine similarity between video \textbf{v} and its corresponding texts \(\textbf{c}_m \in \textbf{C}\) representations:
\begin{equation}
\label{eqn: cosine similarity loss}
\mathrm{sim}(\textbf{v},\textbf{c}_m)=\frac{\langle \textbf{v},\textbf{c}_m\rangle}{\lVert\textbf{v}\rVert \cdot \lVert\textbf{c}_m\rVert}.
\end{equation}

\begin{figure*}[t]
\centering
    \includegraphics[width=0.95\textwidth,height=0.36\textwidth]{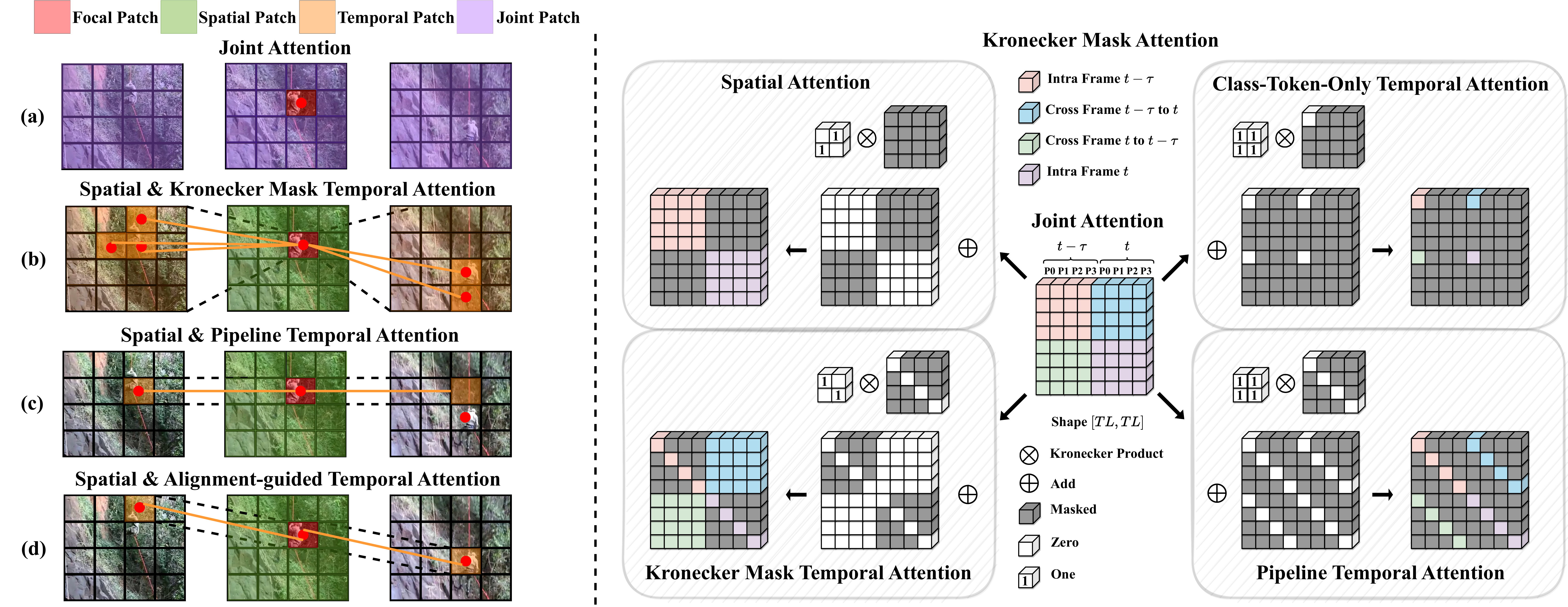}
\caption{(\textbf{Left}) Red indicates the currently focal patch, green patches are visible in spatial attention, orange patches are visible in temporal attention, purple patches are visible in joint attention. (\textbf{Right}) Kronecker mask attention: Several attentions can be seen as employing tailored Kronecker masks for joint attention.}
\label{fig: Kronecker Mask Attention}
\vspace{-0.5cm}
\end{figure*}

\subsection{Kronecker Mask Attention}
\label{sec:kronecker mask attention}
For an image \(\in \mathbb{R}^{H \times W \times 3}\), it is first split into patches and then flattened into a token sequence after patch embedding. The resulting feature shape is \((L, D)\), \(L=\frac{H}{P} \times \frac{W}{P}\), where \(D\) is the hidden dimension. This process is a standard operation in ViT \cite{dosovitskiy2020image}, denoted as Spatial Attention (SA):
\begin{equation}
\label{eqn: spatial attention}
\textbf{Z}_{(L , D)} = \mathrm{Softmax}(\textbf{Q}_{(L , D)}\textbf{K}^\textbf{T}_{(L , D)}/\sqrt{D})\textbf{V}_{(L , D)},
\end{equation}
where \(\textbf{Q}\), \(\textbf{K}\), \(\textbf{V}\) represent the query, key, value matrices, respectively.

For a video \(\in \mathbb{R}^{T \times H \times W \times 3}\), \(T\) is the number of frames, previous studies \cite{arnab2021vivit,bertasius2021space,neimark2021video,guo2021ssan,tong2022videomae,feichtenhofer2022masked} typically employ either joint attention \cite{bertasius2021space,feichtenhofer2022masked} or factorized (divided) attention \cite{bertasius2021space,arnab2021vivit}. Joint attention flattens a video into a longer token sequence, resulting in a feature shape of \((T \times L, D)\). This token interaction mode is illustrated in Fig. \ref{fig: Kronecker Mask Attention} \textbf{Left} (a). Joint attention encounters spatiotemporal homogenization issues, where random shuffling of tokens not affecting (or slightly) the final pooling result. 

In contrast, factorized attention factorize joint attention into spatial attention and temporal attention to avoid spatiotemporal homogenization. They utilize the same spatial attention as Eqn. \ref{eqn: spatial attention}, while the temporal attention varies. Two common temporal attentions are pipeline temporal attention \cite{arnab2021vivit,bertasius2021space} and class-token-only temporal attention \cite{arnab2021vivit,ni2022expanding,wang2021actionclip}. Their feature shapes are \((L, T, D)\) and \((T, D)\), respectively. Pipeline temporal attention has a limited temporal receptive field, which is limited to a fixed time pipeline (tube), hence the term "pipeline temporal attention", as shown in Fig. \ref{fig: Kronecker Mask Attention} \textbf{Left} (c). It has limited scope of capture dynamic information since objects of interest do not always appear in the same 2D location across frames. Although ATA \cite{zhao2022alignment} utilizes alignment techniques to bend the time pipeline to capture dynamic objects, it still has a limited temporal receptive field, in Fig. \ref{fig: Kronecker Mask Attention} \textbf{Left} (d). Similarly, class-token-only temporal attention retains only the class token time pipeline and discard others, which both face limited receptive field and may discard lots of potentially valuable information. It is worth noting that in pipeline temporal attention, mean pooling is necessary across all tokens, otherwise, it is equivalent to class-token-only temporal attention.

To address aforementioned drawbacks, we propose Kronecker Mask Temporal Attention (KMTA). Specifically, we allow each patch (token) at timestamp \(t\) can interact with all other patches (tokens), excluding those sharing the same timestamp \(t\), as illustrated in Fig. \ref{fig: Kronecker Mask Attention}  \textbf{Left} (b). Compared to pipeline temporal attention, KMTA expands the temporal receptive field width of each token. KMTA can be achieved through joint attention incorporated a Kronecker mask, as shown in Fig. \ref{fig: Kronecker Mask Attention}  \textbf{Right} (left down). Additionally, KMTA alleviates the impact of spatiotemporal homogenization due to the presence of the Kronecker mask. The trick for obtaining the Kronecker mask is Kronecker product \(\otimes\):
\begin{equation}
\label{eqn: Kronecker product}
\textbf{A}_{m \times n} \otimes \textbf{B}_{p \times q} =
\begin{bmatrix}
a_{11}\textbf{B} & a_{12}\textbf{B} & \cdots & a_{1n}\textbf{B} \\
a_{21}\textbf{B} & a_{22}\textbf{B} & \cdots & a_{2n}\textbf{B} \\
\vdots & \vdots & \ddots & \vdots \\
a_{m1}\textbf{B} & a_{m2}\textbf{B} & \cdots & a_{mn}\textbf{B}
\end{bmatrix}_{mp \times nq}
\end{equation}
Eqn. \ref{eqn: Kronecker product} is the definition of \(\otimes\), where \(\textbf{A} \in \mathbb{R}^{m \times n}\), \(\textbf{B} \in \mathbb{R}^{p \times q}\). Thus, it is referred to as the Kronecker mask. Kronecker Mask Temporal Attention (KMTA) can be formulated as:
\begin{equation}
\label{eqn: Kronecker mask}
\textbf{M}_{(T \times L,T \times L)} = [\textbf{I}_{(T,T)} \otimes (\textbf{J}_{(L,L)}-\textbf{I}_{(L,L)})]_{1 == -\mathrm{inf}},
\end{equation}
\begin{equation}
\label{eqn: temporal attention}
\textbf{Z}_{(T \times L, D)} = \mathrm{Softmax}(\textbf{Q}_{(T \times L, D)}\textbf{K}^\textbf{T}_{(T \times L, D)}/\sqrt{D}+\textbf{M}_{(T \times L,T \times L)})\textbf{V}_{(T \times L, D)},
\end{equation}
where \(\textbf{I}_{(-,-)}\) is an identity matrix, \(\textbf{J}_{(-,-)}\) is an all-ones matrix, \([\quad]_{1==-\mathrm{inf}}\) means replacing 1 in the matrix with negative infinity (\(-\mathrm{inf}\)), and \(\textbf{M}_{(T \times L,T \times L)}\) is the Kronecker mask. A transformer equipped with KMTA is referred to as a Kronecker Mask Temporal (KMT) transformer.

In fact that both the spatial and temporal attention we mentioned above can be derived by combining a tailored Kronecker mask with joint attention. Therefore, we collectively refer to them as Kronecker Mask Attention, as depicted in Fig. \ref{fig: Kronecker Mask Attention} (\textbf{Right}). The Kronecker mask serves as a prior spatiotemporal heterogeneity inductive bias. Spatial attention allows intra-frame interactions but blocks inter-frame interactions, while Kronecker mask temporal attention allows inter-frame interactions but blocks intra-frame interactions, exhibiting a spatiotemporal structural complementarity.
\begin{wrapfigure}[8]{r}{0.57\textwidth} 
\centering
\vspace{-0.0cm}
\includegraphics[width=0.57\textwidth]{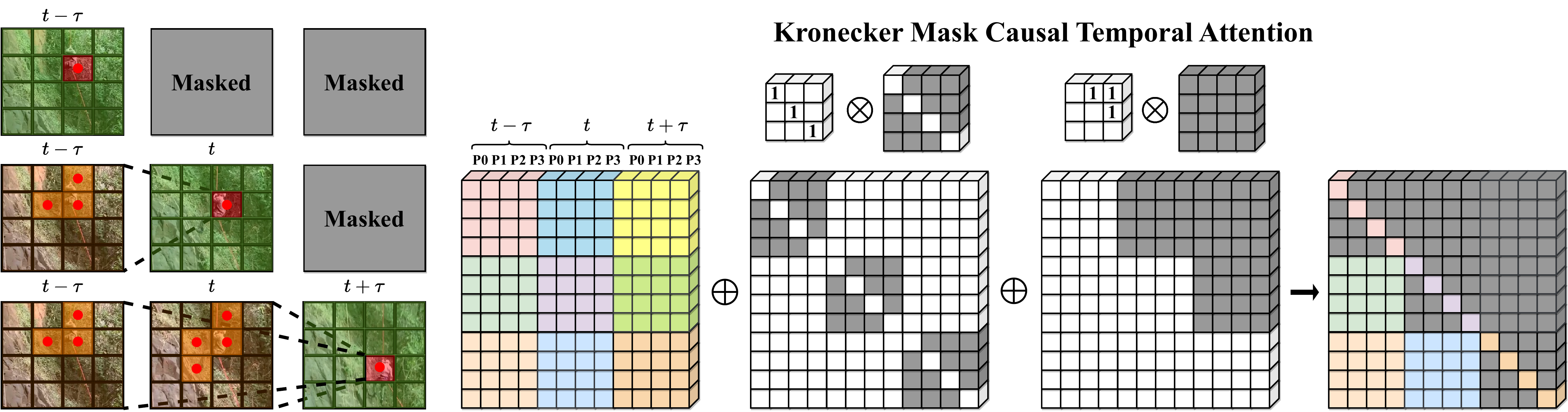} 
\vspace{-0.5cm}
\caption{Kronecker mask causal temporal attention.} 
\label{fig: Kronecker Mask Causal Temporal Attention}
\end{wrapfigure}

Moreover, we design another type of Kronecker mask for temporal modeling, known as Kronecker mask causal temporal attention (KMCTA) that aims to alleviate the low-rank bottleneck, as shown in Fig. \ref{fig: Kronecker Mask Causal Temporal Attention}. The mask \(\tilde{\textbf{M}}_{(T \times L,T \times L)}\) of KMCTA can be formulated as:
\begin{equation}
\label{eqn: Kronecker mask causal}
\tilde{\textbf{M}}_{(T \times L,T \times L)} = [\textbf{I}_{(T,T)} \otimes (\textbf{J}_{(L,L)}-\textbf{I}_{(L,L)}) + (\textbf{U}_{(T,T)}-\textbf{I}_{(T,T)}) \otimes \textbf{J}_{(L,L)}]_{1 == -\mathrm{inf}},
\end{equation}
where \(\textbf{U}_{(-,-)}\) is an upper triangular matrix, and all elements of the upper triangle are 1. A transformer equipped with KMCTA is referred to as a KMCT transformer. KMCTA ensures the causality of time (in the time dimension, KMTA is bidirectional, while KMCTA is unidirectional). In addition, KMCTA always has a full-rank attention matrix, whereas KMTA and joint attention can not guarantee this property. The proof is detailed in \textit{Appendix} \ref{sec: Proof}. Some studies \cite{bhojanapalli2020low,han2023flatten} indicate that it is important to avoid the low-rank bottleneck to improve the representation power of transformer architecture. This problem may become more knotty when transitioning images to videos.


\subsection{Interpretive Prompt}
\label{sec: interpretive prompt}
\begin{wrapfigure}[9]{r}{0.48\textwidth} 
\centering
\vspace{-0.5cm}
\includegraphics[width=0.48\textwidth]{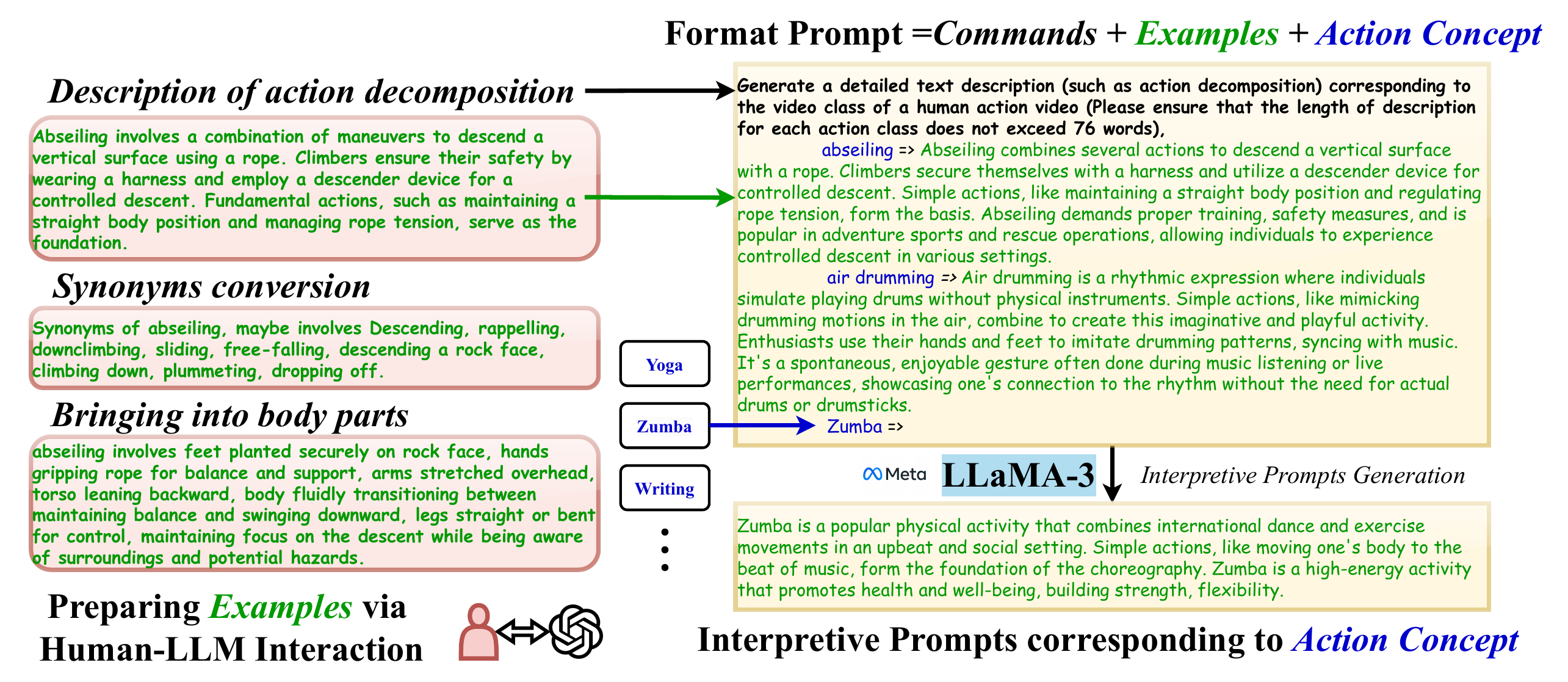}
\vspace{-0.6cm}
\caption{The Interpretive Prompt scheme.} 
\label{fig: Interpretive Prompt}
\end{wrapfigure}
To address the issue 2) in Sec. \ref{sec:intro}, we prompt LLMs to generate interpretive texts that align the text semantics of video-text pairs with the rich text semantics in VLM's pre-training dataset. This design aids the model in understanding abstract verbs. This approach, termed interpretive prompt, designed from the following aspects: 1) Action decomposition: Providing detailed descriptions of actions by decomposing complex actions into simpler, more basic ones. This clarifies the include relationship between complex and basic actions, and help distinguish similar actions, as not all action concepts are of equal status. For example, the action 'playing basketball' may consist of 'running', 'jumping', and 'shooting'. 'Running', 'jumping', and 'hand movements' are considered as basic actions, while 'playing basketball' is a complex action. Similarly, 'dribbling' may also involves 'running', and 'hand movements', besides, it is also a subset of 'playing basketball'. Action decomposition enhances the separability of action concepts in semantic space, helping models understand the relationships between actions. 2) Synonym conversion: Generating synonyms for verbs and phrases that convey the same core concept but with more varied expressions. This improves zero-shot generalization as an action concept may have multiple expressions. When encountering similar action descriptions in the unseen domain, the model demonstrates stronger generalization robustness. 3) Involving body parts: Describing actions based on possible body parts involved, helping the model to localize the region where action occurs.

We leverage ChatGPT \cite{openai2024gpt4technicalreport} and LLaMA-3 \cite{dubey2024llama3herdmodels} to automatically generate action interpretations. Initially, we ask the knowledgeable ChatGPT \cite{openai2024gpt4technicalreport} to provide several examples of text descriptions that align with our expect (based on the aforementioned aspects). Subsequently, we provide the format prompt to LLaMA-3 \cite{dubey2024llama3herdmodels} for text completion, as illustrated in Fig. \ref{fig: Interpretive Prompt}. For example, "\textit{Generate a detailed text description corresponding to the video class...}" --\textit{\textbf{Command}}, accompanied by a few examples like, "\textit{abseiling $\rightarrow$ Abseiling combines several actions...}" --\textit{\textbf{Examples}}, and the action concept that LLM needs to interpret --\textit{\textbf{Action Concept}}. We feed the format prompt to LLaMA-3 multiple times to obtain diverse interpretive prompts of actions. For a given action category, the texts used during training and inference include the original verbs or phrases, template filling prefixes or suffixes, and interpretive prompts. Our interpretive prompts provide texts at the sentence, phrase, and word-levels, enhancing the flexibility of text usage during the inference. More technique details refer to \textit{Appendix} \ref{sec: Interpretive Prompt Details}. 

\subsection{Training and Inference}
\label{sec: Training and Inference}
In each training step, a batch of \(B\) videos is sampled, and \(M\) represents the total number of text descriptions of each verb or phrase. There are a total of \(K\) action categories. The training loss is formulated as follows:
\begin{equation}
\label{eqn: loss}
L=-\frac{1}{N} \sum_m^{\mathrm{sub}\{M\}} \sum_i^{B}   \mathrm{log} \frac{\exp(\mathrm{sim}(\textbf{v}^i,\textbf{c}_m^i)/\tau)}{\sum_k^{K} \exp(\mathrm{sim}(\textbf{v}^i,\textbf{c}_m^k)/\tau)},  
\end{equation}
where, \(\tau\) is the temperature parameter, \(\mathrm{sub}\{M\}\) indicates that we sample a subset of all text descriptions for a single training step. \(\textbf{v}^i\) represents the video belonging to the \(i\)-th category. \(\textbf{c}^i_m\) denotes the descriptions corresponding to the \(i\)-th category. During inference, we take the sum of the similarities between each \(\textbf{v}^i\) and all \(\textbf{c}^i_m\) as the final similarity score \(S\):
\begin{equation}
\label{eqn: inference}
S= \sum_m^{M}   \mathrm{log} \frac{\exp(\mathrm{sim}(\textbf{v}^i,\textbf{c}_m^i)/\tau)}{\sum_k^{K} \exp(\mathrm{sim}(\textbf{v}^i,\textbf{c}_m^k)/\tau)}.  
\end{equation}

\section{Experiments}
\label{sec:experiments}
\subsection{Implementation Details}
\label{sec:experiments setup}
\textbf{Architectures and hyperparameters.} We employ CLIP-B/32, CLIP-B/16, CLIP-L/14 as our backbones and derive corresponding variants: CLAVER-B/32, CLAVER-B/16, CLAVER-L/14, respectively. The frame length settings include 8 and 16. For all CLAVER variants, the number of layers in the KMT/KMCT transformer is equal to one-third of the number of layers in the image encoder. For example, if the image encoder has 12 or 24 layers, then the KMT/KMCT transformer employs 4 or 8 layers. The detailed hyperparameter settings are provided in \textit{Appendix} \ref{sec: Experiment Setting Details}.

\noindent
\textbf{Datasets and metrics.} We evaluate the performance of our method on four benchmarks: Kinetics-400 \cite{kay2017kinetics}, Kientics-600 \cite{carreira2018short}, UCF-101 \cite{soomro2012ucf101}, HMDB-51 \cite{kuehne2011hmdb}. We report the Top-1 and Top-5 accuracy as evaluation metrics.

\begin{table*}[t!]
\centering
\caption{Comparison with state-of-the-art on Kinetics-400. * indicates pretraining with a video-text collection. The FLOPs per view of each method is reported. \textcolor{red}{Red} represents optimal performance. \textbf{Bold} represents optimal performance among CLIP-based methods at the same model scale.}
\label{table: Kinetics-400}
\resizebox{0.92\textwidth}{!}{
\begin{tabular*}{1.52\linewidth}{p{5.3cm}>{\centering\arraybackslash}p{3cm}>{\centering\arraybackslash}p{2cm}>{\centering\arraybackslash}p{2cm}>{\centering\arraybackslash}p{2cm}>{\centering\arraybackslash}p{2cm}>{\centering\arraybackslash}p{2cm}}
\hline
Method & Pretrain & Frames & Top-1 & Top-5 & Views & FLOPs(G) \\
\hline
\multicolumn{7}{c}{\textbf{Without Language}} \\
\hline
Uniformer-B \cite{li2022uniformer} & IN-1k & 32 & 83.0 & 95.4 & 4$\times$3  & 259 \\
TimeSformer-L \cite{bertasius2021space} & IN-21k & 96 & 80.7 & 94.7 & 1$\times$3  & 2380 \\
Mformer-HR \cite{patrick2021keeping} & IN-21k & 16 & 81.1 & 95.2 & 10$\times$3  & 959 \\
ATA \cite{zhao2022alignment} & IN-21k & 32 & 81.9 & 95.5 & 4$\times$3  & 793 \\
Swin-L (384 $\uparrow$) \cite{liu2022video} & IN-21k & 32 & 84.9 & 96.7 & 10$\times$5  & 2107 \\
MViTv2-L (312 $\uparrow$) \cite{li2112improved} & IN-21k & 40 & 86.1 & 97.0 & 5$\times$5  & 2828 \\
ViViT-H/14x2 \cite{arnab2021vivit} & JFT-300M & 32 & 84.9 & 95.8 & 4$\times$3  & 8316 \\
TokenLearner-L/10 \cite{ryoo2021tokenlearner} & JFT-300M & - & 85.4 & 96.3 & 4$\times$3  & 4076 \\
CoVeR \cite{zhang2021co} & JFT-3B & - & 87.2 & - & 1$\times$3  & - \\
\hline
\multicolumn{7}{c}{\textbf{With Language}} \\
\hline
MTV-H \cite{yan2022multiview} & WTS* & 32 & \textcolor{red}{\textbf{89.1}} & \textcolor{red}{\textbf{98.2}} & 4$\times$3  & 3705 \\
\hline
X-CLIP-B/32 \cite{ni2022expanding} & CLIP-400M & 8 & 80.1 & 94.8 & 4$\times$3  & 26 \\
X-CLIP-B/32 \cite{ni2022expanding} & CLIP-400M & 16 & 81.0 & 95.1 & 4$\times$3  & 49 \\
ILA-B/32 \cite{tu2023implicit} & CLIP-400M & 8 & 80.6 & 94.9 & 4$\times$3  & 40 \\
ILA-B/32 \cite{tu2023implicit} & CLIP-400M & 16 & 81.8 & 95.4 & 4$\times$3  & 75 \\
\hline
\rowcolor{gray!25} 
\textbf{CLAVER-B/32 (KMT)} & CLIP-400M & 8 & \textbf{81.5} & \textbf{95.5} & 4$\times$3  & 33 \\
\rowcolor{gray!25} 
\textbf{CLAVER-B/32 (KMT)} & CLIP-400M & 16 & 82.4 & \textbf{95.9} & 4$\times$3  & 64 \\
\rowcolor{gray!25} 
\textbf{CLAVER-B/32 (KMCT)} & CLIP-400M & 8 & 81.4 & \textbf{95.5} & 4$\times$3  & 33 \\
\rowcolor{gray!25} 
\textbf{CLAVER-B/32 (KMCT)} & CLIP-400M & 16 & \textbf{82.6} & \textbf{95.9} & 4$\times$3  & 64 \\
\hline
Action-CLIP-B/16 \cite{wang2021actionclip} & CLIP-400M & 16 & 82.6 & 96.2 & 10$\times$3  & - \\
A6 \cite{ju2022prompting} & CLIP-400M & 16 & 76.9 & 93.5 & -  & - \\
X-CLIP-B/16 \cite{ni2022expanding} & CLIP-400M & 8 & 83.1 & 96.5 & 4$\times$3  & 92 \\
X-CLIP-B/16 \cite{ni2022expanding} & CLIP-400M & 16 & 84.3 & 96.8 & 4$\times$3  & 183 \\
EVL-B/16 \cite{lin2022frozen} & CLIP-400M & 8 & 82.9 & - & -  & 444 \\
EVL-B/16 \cite{lin2022frozen} & CLIP-400M & 16 & 83.6 & - & -  & 888 \\
ViFiCLIP-B/16 \cite{rasheed2023fine} & CLIP-400M & 16 & 83.9 & 96.3 & 4$\times$3  & 281 \\
ASU-B/16 \cite{chen2023video} & CLIP-400M & 8 & 84.1 & 96.3 & 4$\times$3  & 146 \\
ASU-B/16 \cite{chen2023video} & CLIP-400M & 16 & 84.8 & 96.7 & 4$\times$3  & 288 \\
ILA-B/16 \cite{tu2023implicit} & CLIP-400M & 8 & 83.4 & 96.3 & 4$\times$3  & 150 \\
ILA-B/16 \cite{tu2023implicit} & CLIP-400M & 16 & 85.0 & 97.0 & 4$\times$3  & 302 \\
ALT-B/16 \cite{chen2024align} & CLIP-400M & 16 & 85.5 & 96.7 & 3$\times$1  & 1308 \\
\hline
\rowcolor{gray!25} 
\textbf{CLAVER-B/16 (KMT)} & CLIP-400M & 8 & \textbf{84.3} & 96.3 & 4$\times$3  & 122 \\
\rowcolor{gray!25} 
\textbf{CLAVER-B/16 (KMT)} & CLIP-400M & 16 & 85.9 & \textbf{97.3} & 4$\times$3  & 241 \\
\rowcolor{gray!25} 
\textbf{CLAVER-B/16 (KMCT)} & CLIP-400M & 8 & 84.1 & 96.2 & 4$\times$3  & 122 \\
\rowcolor{gray!25} 
\textbf{CLAVER-B/16 (KMCT)} & CLIP-400M & 16 & \textbf{86.0} & 97.2 & 4$\times$3  & 241 \\
\hline
X-CLIP-L/14 \cite{ni2022expanding} & CLIP-400M & 8 & 87.0 & 97.7 & 4$\times$3  & 420 \\
X-CLIP-L/14 (336$\uparrow$) \cite{ni2022expanding} & CLIP-400M & 16 & 87.6 & 97.5 & 4$\times$3  & 1870 \\
EVL-L/14 \cite{lin2022frozen} & CLIP-400M & 8 & 86.3 & - & -  & 2022 \\
EVL-L/14 (336$\uparrow$) \cite{lin2022frozen}  & CLIP-400M & 32 & 87.7 & - & -  & 18196 \\
ASU-L/14 \cite{chen2023video} & CLIP-400M & 8 & 87.8 & 97.8 & 4$\times$3  & 660 \\
ASU-L/14 (336$\uparrow$) \cite{chen2023video} & CLIP-400M & 16 & 88.3 & 98.0 & 4$\times$3  & 3084 \\
ILA-L/14 \cite{tu2023implicit} & CLIP-400M & 8 & 87.6 & 97.8 & 4$\times$3  & 647 \\
ILA-L/14 (336$\uparrow$) \cite{tu2023implicit}  & CLIP-400M & 16 & 88.1 & 97.8 & 4$\times$3  & 3130 \\
ALT-B/14 \cite{chen2024align} & CLIP-400M & 16 & 87.8 & 97.7 & 3$\times$1  & 4947 \\
\hline
\rowcolor{gray!25} 
\textbf{CLAVER-L/14 (KMT)} & CLIP-400M & 8 & \textbf{88.1} & 97.7 & 4$\times$3  & 558 \\
\rowcolor{gray!25} 
\textbf{CLAVER-L/14(336$\uparrow$) (KMT)} & CLIP-400M & 16 & 88.8 & \textbf{98.1} & 4$\times$3  & 2488 \\
\rowcolor{gray!25} 
\textbf{CLAVER-L/14 (KMCT)} & CLIP-400M & 8 & 87.9 & 97.7 & 4$\times$3  & 558 \\
\rowcolor{gray!25} 
\textbf{CLAVER-L/14(336$\uparrow$) (KMCT)} & CLIP-400M & 16 & \textbf{88.9} & 98.0 & 4$\times$3  & 2488 \\
\hline
\end{tabular*}
}
\vspace{-0.3cm}
\end{table*}

 \subsection{Comparison Results}
\label{sec:Comparison Results}
\textbf{Fully-supervised Experiments.} We conducted fully supervised experiments on Kinetics-400 and Kinetics-600, respectively. In Tab. \ref{table: Kinetics-400}, we employ three variant models, CLAVER-B/32, CLAVER-B/16, and CLAVER-L/14, and sample 8 or 16 frames (\(8f\),\(16f\)) with a sparse sampling for each model, employ KMT/KMCT, respectively. CLAVER-B/16\(_{8f}\) (KMT/KMCT) and CLAVER-B/16\(_{16f}\) (KMT/KMCT) surpass several methods \cite{bertasius2021space,patrick2021keeping,zhao2022alignment,li2022uniformer} pretrained on ImageNet-1k/21k \cite{deng2009imagenet} with shorter frames. CLAVER-B/16\(_{16f}\) (KMT) outperforms Swin-L (384$\uparrow$) \cite{liu2022video} by 1.0\%, and is slightly lower than MViTv2-L (312$\uparrow$) \cite{li2112improved}, as lower resolution, shorter frames and fewer views. CLAVER-B/14\(_{8f}\) (KMT/KMCT) outperforms some methods \cite{arnab2021vivit,ryoo2021tokenlearner,zhang2021co} pretrained on JFT-300M/JFT-3B. CLAVER-B/14\(_{16f}\) (KMT) outperforms CoVeR \cite{zhang2021co} by 1.6\%, however, inferior to MTV-H \cite{yan2022multiview}, as it utilizes WTS*, which contains 70M video-text pairs with about 17B images, much larger than CLIP-400M. Compared to those approaches based on CLIP \cite{radford2021learning}, under configuration ViT-B/32\(_{8f}\) and ViT-B/32\(_{16f}\), CLAVER (KMT/KMCT) surpasses X-CLIP \cite{ni2022expanding} and ILA \cite{tu2023implicit}. Under configuration ViT-B/16\(_{8f,16f}\) and ViT-L/14\(_{8f,16f}\), CLAVER (KMT/KMCT) exceeds most methods under the same configuration. In Tab. \ref{table: Kinetics-600}, CLAVER-B/16\(_{8f}\) (KMT) achieves higher performance compared to ViViT-H/14x2 \cite{arnab2021vivit}, MViT-B-24 \cite{fan2021multiscale}. Our method has lower performance than these methods \cite{ryoo2021tokenlearner,zhang2021co,yan2022multiview,yuan2021florence,liu2022video}, as they use longer frames, or more data, or higher resolutions. In addition, CLAVER-B/16\(_{8f}\) (KMT/KMCT) outperforms ASU-B/16\(_{8f}\) \cite{chen2023video} and X-CLIP-B/16\(_{8f}\) \cite{ni2022expanding} at the same scale.

\begin{table}[t!]
\centering
\begin{minipage}[t]{0.48\linewidth}
\centering
\caption{Comparison with state-of-the-art on Kinetics-600.}
\vspace{-0.3cm}
\label{table: Kinetics-600}
\resizebox{1.0\linewidth}{!}{
\begin{tabular}{ccccccccccc}
\hline
\textbf{Method} & \textbf{Pretrain} & \textbf{Frames} & \textbf{Top-1} & \textbf{Top-5} & \textbf{Views}\\
\hline
MViT-B-24 \cite{fan2021multiscale} & - & 32 & 83.8 & 96.3 & 5$\times$1 \\
Swin-L (384 $\uparrow$) \cite{liu2022video} & JFT-300M & 32 & 85.8 & 96.5 & 4$\times$3  \\
ViViT-H/14x2 \cite{arnab2021vivit}  & JFT-300M & 32 & 85.8 & 96.5 & 4$\times$3  \\
TokenLearner-L/10 \cite{ryoo2021tokenlearner} & JFT-300M & - & 86.3 & 97.0 & 4$\times$3 \\
CoVeR \cite{zhang2021co} & JFT-3B & 32 & 87.9 & - & 4$\times$3\\
MTV-H \cite{yan2022multiview} & WTS* & 32 & 89.6 & 98.3 & 4$\times$3 \\
Florence (384 $\uparrow$) \cite{yuan2021florence} & FLD-900M & - & 87.8 & - & 1$\times$3 \\
X-CLIP-B/16 \cite{ni2022expanding} & CLIP-400M & 8 & 85.3 & 97.1 & 4$\times$3 \\
ASU-B/16 \cite{chen2023video} & CLIP-400M & 8 & 85.7 & - & 4$\times$3  \\
\hline
\rowcolor{gray!25} 
\textbf{CLAVER-B/16 (KMT)} & CLIP-400M & 8 & 85.9 & 97.3 & 4$\times$3 \\
\hline
\end{tabular}
}
\end{minipage}
\hspace{0.05cm}
\begin{minipage}[t]{0.48\linewidth}
\centering
\caption{Few-shot performance on HMDB-51 and UCF-101.}
\vspace{-0.3cm}
\label{table: few-shot HMDB-51 UCF-101}
\resizebox{1.0\linewidth}{!}{
\begin{tabular}{cccccccccc}
\hline
\multirow{2}{*}{\textbf{Method}} & \multicolumn{4}{c}{\textbf{HMDB-51}} &  & \multicolumn{4}{c}{\textbf{UCF-101}} \\
\cmidrule(r){2-5} \cmidrule(r){7-10}
& K=2 & K=4 & K=8 & K=16 &  & K=2 & K=4 & K=8 & K=16 \\
\hline
TSM \cite{lin2019tsm} & 17.5 & 20.9 & 18.4 & 31.0 &   & 25.3 & 47.0 & 64.4 & 61.0 \\
TimeSformer \cite{bertasius2021space} & 19.6 & 40.6 & 49.4 & 55.4 &  & 48.5 & 75.6 & 83.7 & 89.4 \\
Swin-B \cite{liu2022video} & 20.9 & 41.3 & 47.9 & 56.1 &  & 53.3 & 74.1 & 85.8 & 88.7 \\
Action-CLIP \cite{wang2021actionclip} & 55.0 & 56.0 & 58.0 & - &  & 80.0 & 85.0 & 89.0 & - \\
X-CLIP \cite{ni2022expanding} & 53.0 & 57.3 & 62.8 & 64.0 &  & 76.4 & 83.4 & 88.3 & 91.4 \\
X-Florence \cite{ni2022expanding} & 51.6 & 57.8 & 64.1 & 64.2 &  & 84.0 & 88.5 & 92.5 & 94.8 \\
MAXI \cite{lin2023match} & 58.0 & 60.1 & 65.0 & 66.5 & & 86.8 & 89.3 & 92.4 & 93.5 \\
ASU \cite{chen2023video} & \textbf{60.1} & 63.8 & 67.2 & 70.8 & & \textbf{91.4} & \textbf{94.6} & 96.0 & 97.2 \\
\hline
\rowcolor{gray!25} 
\textbf{CLAVER (KMT)}&  58.6 &  \textbf{63.9} & \textbf{68.0} & \textbf{72.5} &  &  89.7 & 92.9 & \textbf{96.1} & \textbf{98.0} \\
\hline
\end{tabular}
}
\end{minipage}
\vspace{-0.1cm}
\end{table}

\begin{table}[t!]
\vspace{-0.1cm}
 \label{table: zero-shot}
    \begin{minipage}[t]{0.48\linewidth}
    \centering
\caption{Zero-shot on HMDB-51 and UCF-101.}
\vspace{-0.3cm}
\label{table: zero-shot HMDB-51 UCF-101}
\resizebox{0.82\textwidth}{!}{
\begin{tabular}{ccc}
\hline
\textbf{Method} & \textbf{HMDB-51} & \textbf{UCF-101} \\
\hline
MTE \cite{xu2016multi} & 19.7 $\pm$ 1.6 & 15.8 $\pm$ 1.3 \\
ASR \cite{wang2017alternative} & 21.8 $\pm$ 0.9 & 24.4 $\pm$ 1.0 \\
ZSECOC \cite{qin2017zero} & 22.6 $\pm$ 1.2 & 15.1 $\pm$ 1.7 \\
UR \cite{zhu2018towards} & 24.4 $\pm$ 1.6 & 17.5 $\pm$ 1.6 \\
TS-GCN \cite{gao2019know} & 23.2 $\pm$ 3.0 & 34.2 $\pm$ 3.1 \\
E2E \cite{brattoli2020rethinking} & 32.7 & 48 \\
ER-ZSAR \cite{chen2021elaborative} & 35.3 $\pm$ 4.6 & 51.8 $\pm$ 2.9 \\
Action-CLIP \cite{wang2021actionclip} & 40.8 $\pm$ 5.4  & 58.3 $\pm$ 3.4 \\
X-CLIP \cite{ni2022expanding} & 44.6 $\pm$ 5.2  & 72.0 $\pm$ 2.3 \\
ASU \cite{chen2023video} & 48.1 $\pm$ 2.8  & 75.0 $\pm$ 2.3 \\
MAXI \cite{lin2023match} & 51.2 $\pm$ 1.1  & 75.2 $\pm$ 0.9 \\
OST \cite{chen2024ost} & 52.9 $\pm$ 0.9  & 75.3 $\pm$ 2.1 \\
\hline
\rowcolor{gray!25} 
\textbf{CLAVER (KMT)} & \textbf{54.1 $\pm$ 2.4} & \textbf{78.6 $\pm$ 1.7}\\
\hline
\end{tabular}} 
    \end{minipage}
    \hspace{0.2cm}
    \begin{minipage}[t]{0.48\linewidth} 
        \centering
\caption{Zero-shot on Kinetics-600.}
\vspace{-0.1cm}
\label{table: zero-shot Kinetics600}
\resizebox{0.87\textwidth}{!}{
\begin{tabular}{ccc}
\hline
\textbf{Method} & \textbf{Top-1} & \textbf{Top-5} \\
\hline
DEVISE \cite{frome2013devise} & 23.8 $\pm$ 0.3 & 51.0 $\pm$ 0.6 \\
ALE \cite{akata2015label} & 23.4 $\pm$ 0.8 & 50.3 $\pm$ 1.4 \\
SJE \cite{akata2015evaluation} & 22.3 $\pm$ 0.6 & 48.2 $\pm$ 0.4 \\
ESZSL \cite{romera2015embarrassingly} & 22.9 $\pm$ 1.2 & 48.3 $\pm$ 0.8 \\
DEM \cite{zhang2017learning} & 23.6 $\pm$ 0.7 & 49.5 $\pm$ 0.4 \\
GCN \cite{ghosh2020all} & 22.3 $\pm$ 0.6 & 49.7 $\pm$ 0.6 \\
ER-ZSAR \cite{chen2021elaborative} & 42.1 $\pm$ 1.4 & 73.1 $\pm$ 0.3 \\
X-CLIP \cite{ni2022expanding}& 65.2 $\pm$ 0.4 & 86.1 $\pm$ 0.8 \\
ASU \cite{chen2023video} & 67.6 $\pm$ 0.2& 87.2 $\pm$ 0.3 \\
MAXI \cite{lin2023match} & 70.9 $\pm$ 1.2  & 92.1 $\pm$ 0.5 \\
OST \cite{chen2024ost} & 70.5 $\pm$ 0.7  & 92.1 $\pm$ 0.3 \\
\hline
\rowcolor{gray!25} 
\textbf{CLAVER (KMT)}  & \textbf{73.8 $\pm$ 0.6}& \textbf{93.1 $\pm$ 0.6}\\
\hline
\end{tabular}}
    \end{minipage}
    \vspace{-0.5cm}
\end{table}


\textbf{Few-shot Experiments.} We pretrain CLAVER-B/16\(_{32f}\) (KMT) (\(32f\) means with 32 frames) on Kinetics-400, and then perform few-shot transfer of 2, 4, 8, 16 samples on HMDB-51 and UCF-101. Tab. \ref{table: few-shot HMDB-51 UCF-101} depicts the results of few-shot experiments. CLAVER is comparable wtih ASU \cite{chen2023video}, and consistently surpasses the X-CLIP \cite{ni2022expanding} and MAXI \cite{lin2023match} across all K ranges. Additionally, CLAVER significantly outperforms other previous methods like \cite{lin2019tsm,bertasius2021space,liu2022video}. More details about the evaluation protocols are provided in the \textit{Appendix} \ref{sec: Experiment Setting Details}.

\textbf{Zero-shot Experiments.} We also pretrain CLAVER-B/16\(_{32f}\) (KMT) on Kinetics400 for zero-shot. As shown in Tab. \ref{table: zero-shot HMDB-51 UCF-101}, on HMDB-51 \cite{kuehne2011hmdb} and UCF-101 \cite{soomro2012ucf101} benchmarks, CLAVER surpasses OST \cite{chen2024ost} , ASU \cite{chen2023video} and X-CLIP \cite{ni2022expanding} under the same configuration, and far outperforms other previous methods. Additionally, in Tab. \ref{table: zero-shot Kinetics600}, on Kinetics600 \cite{carreira2018short} benchmark, CLAVER outperforms OST \cite{chen2024ost}, MAXI \cite{lin2023match} and all other methods. More details about the evaluation protocols are provided in the \textit{Appendix} \ref{sec: Experiment Setting Details}.

\subsection{Ablation Study}
\label{sec:ablation study}
\textbf{Components ablation studies.} We performed ablation studies to evaluate the effects of each component under the CLAVER-B/32\(_{8f}\) (KMT) configuration. The results are shown in Tab. \ref{table: ablation analysis of each component}. Our baseline, denoted as CLIP-Mean, implements temporal mean pooling for CLIP. By equipping the CLIP with a KMT transformer at 1/3 scale, the Top-1 accuracy increases by 3.5\%. We only introduce interpretive prompt, the performance increases by 1\%. When we further incorporate both of them, CLAVER surpasses the CLIP-Mean by 4.1\%. In addition, we test the effect of the number of KMT transformer layers. With only one layer, the performance improvement is minimal. Increasing the number of layers to 1/6 scale results in further performance gains, and at 1/3 scale, we observe significant improvement.

\begin{table}[t!]
 \begin{minipage}[t]{0.5\textwidth}
        \centering
        \caption{Ablation of each component.}
        \vspace{-0.2cm}
        \label{table: ablation analysis of each component}
 \resizebox{0.99\textwidth}{!}{   
        \begin{tabular}{cc}
            \toprule
\textbf{Components}& \textbf{Top-1 (\%)}\\
\hline
Baseline (CLIP-Mean) & 77.4\\
Baseline + KMTA 1 layer & 78.3\\
Baseline + KMTA 1/6 scale & 79.6\\
Baseline + KMTA 1/3 scale & 80.9\\
Baseline + Interpretive Prompt & 78.4 \\
Baseline + KMTA 1/3 scale + Interpretive Prompt & \textbf{81.5} \\
\bottomrule
        \end{tabular}
}
    \end{minipage}
    \hspace{0.05cm}
    \begin{minipage}[t]{0.5\textwidth}
        \centering
 \caption{Comparison of temporal attentions.}
        \vspace{-0.2cm}
     \label{table: comparison of different temporal attention}
            \resizebox{1.0\textwidth}{!}{   
        \begin{tabular}{ccc}
            \toprule
\textbf{Temporal Modeling} & \textbf{Top-1 (\%)} & \textbf{Top-5 (\%)}\\
\hline
Baseline (Mean Pooling)& 78.4& 94.3\\
Class-Token-Only& 78.9& 94.3\\
Pipeline Temporal Attention& 79.4& 94.4\\
Joint Attention& 80.1& 94.9\\
Kronecker Mask Causal Temporal Attention& 81.4& \textbf{95.5}\\
Kronecker Mask Temporal Attention& \textbf{81.5}& \textbf{95.5}\\
\bottomrule
        \end{tabular}
}
    \end{minipage}
\vspace{-0.2cm}
\end{table}


\begin{table}[t]
        \centering
        \caption{The impact of patch size and frame length on joint attention (JA), KMTA and KMCTA. Conducting on HMDB-51 and UCF-101.}
        \vspace{-0.2cm}
     \label{table: Comparison of JA, KMTA and KMCTA}
            \resizebox{0.85\textwidth}{!}{   
\begin{tabular}{ccccccc}
            \toprule
\multirow{2}{*}{\textbf{HMDB-51}, \textbf{UCF-101}}&  \multicolumn{3}{c}{patch size = 32 (ViT-B/32)}& \multicolumn{3}{c}{patch size = 16 (ViT-B/16)} \\
\cmidrule(r){2-7}
& JA (\%)& KMTA (\%)& KMCTA (\%)& JA (\%)& KMTA (\%)& KMCTA (\%)\\
\hline
frame length = 8& 67.9, 93.3& 68.8, 94.1& 68.7, 93.9& 72.2, 96.2& 73.2, 96.3& 72.2, 96.1\\
frame length = 16& 68.0, 93.6&69.1, 94.2& 69.4, 94.8& 72.1, 96.3& 72.3, 96.6& 72.8, 96.4\\
\bottomrule
\end{tabular}
}
\vspace{-0.2cm}
\end{table}

\begin{figure*}[t!]
    \centering
\includegraphics[width=0.97\textwidth,height=0.12\textwidth]{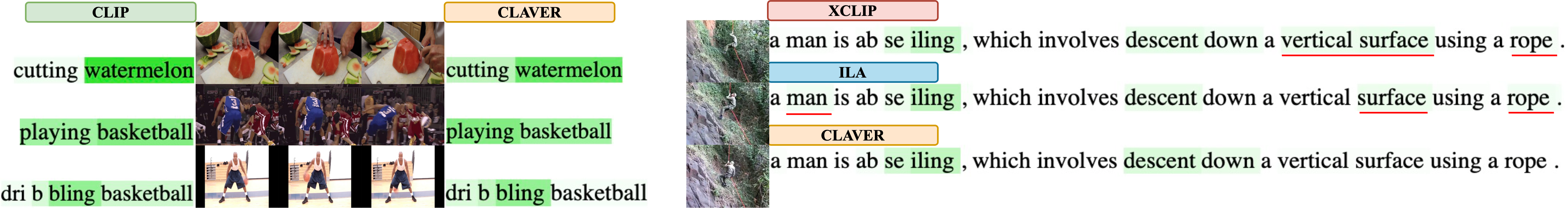}
    \caption{Word importance of CLIP, X-CLIP, ILA and CLAVER. Darker color, higher importance.}
\vspace{-0.3cm}
    \label{fig: Word Importance}
\vspace{-0.1cm}
\end{figure*}

\textbf{Comparison of different temporal attentions and low-rank bottleneck issue.} Tab. \ref{table: comparison of different temporal attention} compares the performance of different temporal modeling methods mentioned in Sec. \ref{sec:kronecker mask attention}. We find that class-token-only temporal attention and pipeline temporal attention have inferior performance. KMTA and KMCTA outperform joint attention. The top-1 (\%) of KMTA slightly surpasses KMCTA by 0.1\%. Then, we conduct further experiments on HMDB-51 and UCF101 that observing the impact of the low-rank bottleneck issue on them. We increasing the length of token sequences by reducing patch size or increasing frame length, and observe the effects on them, as shown in Tab. \ref{table: Comparison of JA, KMTA and KMCTA}. When reducing the patch size, however, we do not observe the low-rank bottleneck. Joint attention, KMTA and KMCTA achieve better performance due to more fine-grained features as the smaller patch size results in each token representing smaller local region. In contrast, when we increase the frame length, KMCTA's performance is optimal in most configurations, and only the performance of KMCTA can steadily improve. The performance improvement of joint attention and KMTA is limited. Meanwhile, in Tab. \ref{table: Kinetics-400}, with the increase of frame length under the same backbone, the performance increase of KMCTA is also greater than that of KMTA. This indicates that KMCTA has a more significant advantage with longer frame length.

\begin{figure*}[t]
\centering
\includegraphics[width=0.98\textwidth,height=0.32\textwidth]{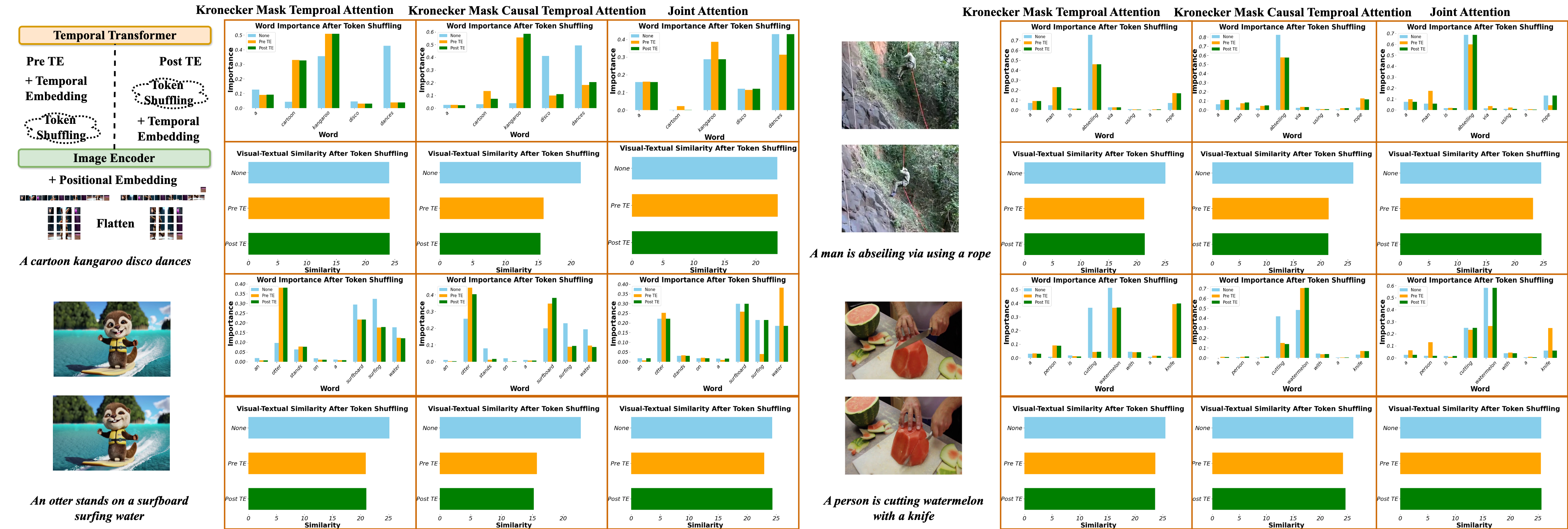}
\caption{Spatiotemporal Homogenization. (\textbf{Upper Left}) Token shuffling. (\textbf{Vertical bar chart}) Word importance refers to the degree of correlation between each word in a sentence description and the semantics of the video content, while (\textbf{Horizontal bar chart}) Similarity refers to the similarity between visual and textual representations.}
\vspace{-0.3cm}
    \label{fig: Spatiotemporal Homogenizations}
\vspace{-0.4cm}
\end{figure*}


\section{Visualization and Analysis}
\label{sec: visualization analysis}
We employ \cite{chefer2021generic} for bi-modal visualization and show the explainability of visual-textual attentions. 

\textbf{Word importance}. We visualize word importance for CLIP, X-CLIP, ILA and CLAVER in Fig. \ref{fig: Word Importance}. Observations indicate that CLIP tends to focus on nouns, whereas CLAVER prefers to verbs. Compared to previous works, CLAVER is more inclined towards verb concept, while X-CLIP \cite{ni2022expanding} and ILA \cite{tu2023implicit} show a slight inclination towards nouns. These findings indicate the effectiveness of interpretive prompts for nouns concept to verbs concept transition.

\textbf{Spatiotemporal homogenization study}. We define spatiotemporal homogenization as a phenomenon where random token shuffling has little impact on the semantics of the visual branch, which is illogical. It aims to illustrate the interpretability behind the performance of spatiotemporal modeling. In Fig. \ref{fig: Spatiotemporal Homogenizations} (\textbf{Upper Left}), we illustrate the token shuffling. We denote token shuffling before adding time embedding as PreTE shuffling, token shuffling after adding time embedding as PostTE shuffling, and None represents no shuffle. Fig. \ref{fig: Spatiotemporal Homogenizations} shows changes in word importance and similarity following both PreTE and PostTE shuffling. For joint attention, we observe that PreTE shuffling marginally affects similarity and word importance, while PostTE shuffling does not affect similarity score and word importance. Regarding KMTA, both PreTE and PostTE shuffling lead to changes in the similarity and word importance. Meanwhile, KMCTA is profoundly affected by both PreTE and PostTE shuffling, which result in lower similarity and disturbance of word importance. This phenomenon suggest that KMTA and KMCTA possess varying degrees ability in mitigating spatiotemporal homogenization and the Kronecker mask serves as a natural inductive bias for spatiotemporal structural heterogeneity. It also demonstrates that equipping learnable position/time encoding is not insufficient to alleviate spatiotemporal homogenization. Please refer to \textit{Appendix} \ref{sec: Spatiotemporal Homogenization Study} for more and explanations and visualizations.

\begin{figure*}[t!]
    \centering
    \includegraphics[width=0.9\textwidth,height=0.36\textwidth]{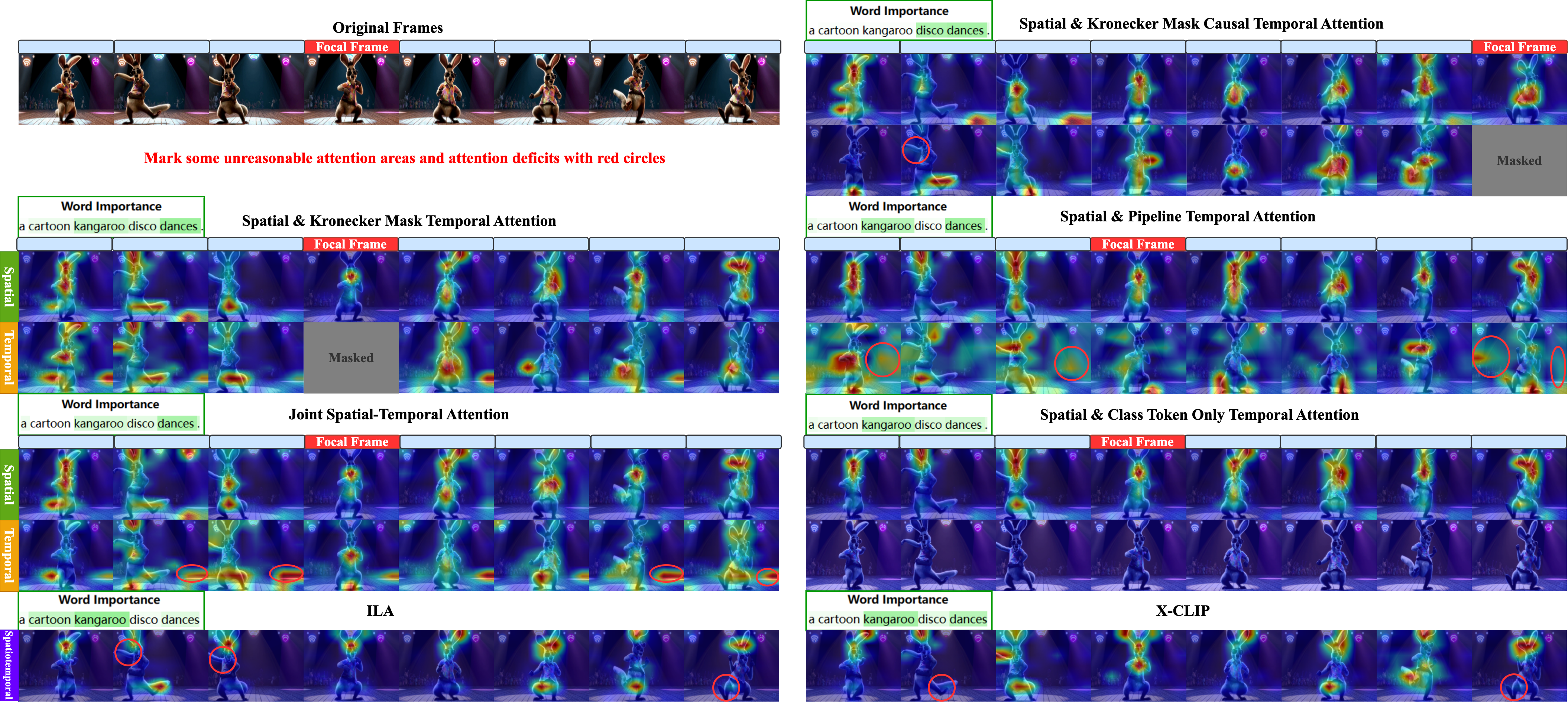}
    \caption{Visualizing spatiotemporal attention maps of different spatio-temporal modeling.}
\vspace{-0.3cm}
    \label{fig: Spatiotemporal attention maps}
\vspace{-0.5cm}
\end{figure*}

\textbf{Visualization of spatiotemporal attention map}. Fig. \ref{fig: Spatiotemporal attention maps} presents various spatiotemporal attention maps. X-CLIP \cite{ni2022expanding} and ILA \cite{tu2023implicit} adopt factorized attention structures resulting in intertwined spatiotemporal attention maps. However, CLAVER employs a factorized encoder structure, allowing for the separation of spatial and temporal attention maps. Notably, class-token-only temporal attention discards other time tubes so that its temporal attention maps appear vacant. KMTA and KMCT have more reasonable spatial (focuses on the executor of the action) and temporal (focuses on areas where the action occurs) attention maps, while ILA and X-CLIP ignore some action dense areas. Moreover, joint attention and pipeline temporal attention will be attracted by some irrelevant backgrounds.

\section{Conclusion}
\label{sec:conclusion}
In this work, we present the \textbf{C}ontrastive \textbf{L}anguage-\textbf{A}ction \textbf{V}ideo Learn\textbf{er} (\textbf{CLAVER}) to shift from the alignment of visual objects and nouns in CLIP to the alignment of action behaviors and verbs. We propose Kronecker mask temporal attention and Kronecker mask causal temporal attention for temporal modeling. Interpretive prompts are employed to transition the focus on nouns to verbs. Extensive experiments under different evaluation settings demonstrate the effectiveness of our method.


\section{Acknowledgement}
\label{sec:acknowledgement}
The numerical calculations in this paper have been done on the supercomputing system in the Supercomputing Center of University of Science and Technology of China. This work was supported by the National Science Foundation of China, under Grant No. 62171425. This work was supported by National Natural Science Foundation of China under Grant 62306061. This work was supported by Guangdong Basic and Applied Basic Research Foundation (Grant No. 2023A1515140037). This work was supported by Guangdong Research Team for Communication and Sensing Integrated with Intelligent Computing (Project No. 2024KCXTD047).

\bibliography{iclr2025_conference}

\begin{thebibliography}{99}
\providecommand{\natexlab}[1]{#1}
\providecommand{\url}[1]{\texttt{#1}}
\expandafter\ifx\csname urlstyle\endcsname\relax
  \providecommand{\doi}[1]{doi: #1}\else
  \providecommand{\doi}{doi: \begingroup \urlstyle{rm}\Url}\fi

\bibitem[Abhimanyu~Dubey et~al.(2024)Abhimanyu~Dubey, Pandey, and et~al.]{dubey2024llama3herdmodels}
Abhinav~Jauhri Abhimanyu~Dubey, Abhinav Pandey, and et~al.
\newblock The llama 3 herd of models, 2024.
\newblock URL \url{https://arxiv.org/abs/2407.21783}.

\bibitem[Akata et~al.(2015{\natexlab{a}})Akata, Perronnin, Harchaoui, and Schmid]{akata2015label}
Zeynep Akata, Florent Perronnin, Zaid Harchaoui, and Cordelia Schmid.
\newblock Label-embedding for image classification.
\newblock \emph{IEEE transactions on pattern analysis and machine intelligence}, 38\penalty0 (7):\penalty0 1425--1438, 2015{\natexlab{a}}.

\bibitem[Akata et~al.(2015{\natexlab{b}})Akata, Reed, Walter, Lee, and Schiele]{akata2015evaluation}
Zeynep Akata, Scott Reed, Daniel Walter, Honglak Lee, and Bernt Schiele.
\newblock Evaluation of output embeddings for fine-grained image classification.
\newblock In \emph{Proceedings of the IEEE conference on computer vision and pattern recognition}, pp.\  2927--2936, 2015{\natexlab{b}}.

\bibitem[Arnab et~al.(2021)Arnab, Dehghani, Heigold, Sun, Lu{\v{c}}i{\'c}, and Schmid]{arnab2021vivit}
Anurag Arnab, Mostafa Dehghani, Georg Heigold, Chen Sun, Mario Lu{\v{c}}i{\'c}, and Cordelia Schmid.
\newblock Vivit: A video vision transformer.
\newblock In \emph{Proceedings of the IEEE/CVF international conference on computer vision}, pp.\  6836--6846, 2021.

\bibitem[Bertasius et~al.(2021)Bertasius, Wang, and Torresani]{bertasius2021space}
Gedas Bertasius, Heng Wang, and Lorenzo Torresani.
\newblock Is space-time attention all you need for video understanding?
\newblock In \emph{ICML}, volume~2, pp.\ ~4, 2021.

\bibitem[Bhojanapalli et~al.(2020)Bhojanapalli, Yun, Rawat, Reddi, and Kumar]{bhojanapalli2020low}
Srinadh Bhojanapalli, Chulhee Yun, Ankit~Singh Rawat, Sashank Reddi, and Sanjiv Kumar.
\newblock Low-rank bottleneck in multi-head attention models.
\newblock In \emph{International conference on machine learning}, pp.\  864--873. PMLR, 2020.

\bibitem[Brattoli et~al.(2020)Brattoli, Tighe, Zhdanov, Perona, and Chalupka]{brattoli2020rethinking}
Biagio Brattoli, Joseph Tighe, Fedor Zhdanov, Pietro Perona, and Krzysztof Chalupka.
\newblock Rethinking zero-shot video classification: End-to-end training for realistic applications.
\newblock In \emph{Proceedings of the IEEE/CVF Conference on Computer Vision and Pattern Recognition}, pp.\  4613--4623, 2020.

\bibitem[Brooks et~al.(2024)Brooks, Peebles, Holmes, DePue, Guo, Jing, Schnurr, Taylor, Luhman, Luhman, Ng, Wang, and Ramesh]{videoworldsimulators2024}
Tim Brooks, Bill Peebles, Connor Holmes, Will DePue, Yufei Guo, Li~Jing, David Schnurr, Joe Taylor, Troy Luhman, Eric Luhman, Clarence Ng, Ricky Wang, and Aditya Ramesh.
\newblock Video generation models as world simulators.
\newblock 2024.
\newblock URL \url{https://openai.com/research/video-generation-models-as-world-simulators}.

\bibitem[Brown et~al.(2020)Brown, Mann, Ryder, Subbiah, Kaplan, Dhariwal, Neelakantan, Shyam, Sastry, Askell, et~al.]{brown2020language}
Tom Brown, Benjamin Mann, Nick Ryder, Melanie Subbiah, Jared~D Kaplan, Prafulla Dhariwal, Arvind Neelakantan, Pranav Shyam, Girish Sastry, Amanda Askell, et~al.
\newblock Language models are few-shot learners.
\newblock \emph{Advances in neural information processing systems}, 33:\penalty0 1877--1901, 2020.

\bibitem[Carreira \& Zisserman(2017)Carreira and Zisserman]{carreira2017quo}
Joao Carreira and Andrew Zisserman.
\newblock Quo vadis, action recognition? a new model and the kinetics dataset.
\newblock In \emph{proceedings of the IEEE Conference on Computer Vision and Pattern Recognition}, pp.\  6299--6308, 2017.

\bibitem[Carreira et~al.(2018)Carreira, Noland, Banki-Horvath, Hillier, and Zisserman]{carreira2018short}
Joao Carreira, Eric Noland, Andras Banki-Horvath, Chloe Hillier, and Andrew Zisserman.
\newblock A short note about kinetics-600.
\newblock \emph{arXiv preprint arXiv:1808.01340}, 2018.

\bibitem[Chefer et~al.(2021)Chefer, Gur, and Wolf]{chefer2021generic}
Hila Chefer, Shir Gur, and Lior Wolf.
\newblock Generic attention-model explainability for interpreting bi-modal and encoder-decoder transformers.
\newblock In \emph{Proceedings of the IEEE/CVF International Conference on Computer Vision}, pp.\  397--406, 2021.

\bibitem[Chen \& Huang(2021)Chen and Huang]{chen2021elaborative}
Shizhe Chen and Dong Huang.
\newblock Elaborative rehearsal for zero-shot action recognition.
\newblock In \emph{Proceedings of the IEEE/CVF International Conference on Computer Vision}, pp.\  13638--13647, 2021.

\bibitem[Chen et~al.(2024{\natexlab{a}})Chen, Yu, Yang, Li, Sun, and Chen]{chen2024ost}
Tongjia Chen, Hongshan Yu, Zhengeng Yang, Zechuan Li, Wei Sun, and Chen Chen.
\newblock Ost: Refining text knowledge with optimal spatio-temporal descriptor for general video recognition.
\newblock In \emph{Proceedings of the IEEE/CVF Conference on Computer Vision and Pattern Recognition}, pp.\  18888--18898, 2024{\natexlab{a}}.

\bibitem[Chen et~al.(2023)Chen, Chen, Liu, Li, and Peng]{chen2023video}
Yifei Chen, Dapeng Chen, Ruijin Liu, Hao Li, and Wei Peng.
\newblock Video action recognition with attentive semantic units.
\newblock In \emph{Proceedings of the IEEE/CVF International Conference on Computer Vision}, pp.\  10170--10180, 2023.

\bibitem[Chen et~al.(2024{\natexlab{b}})Chen, Chen, Liu, Zhou, Xue, and Peng]{chen2024align}
Yifei Chen, Dapeng Chen, Ruijin Liu, Sai Zhou, Wenyuan Xue, and Wei Peng.
\newblock Align before adapt: Leveraging entity-to-region alignments for generalizable video action recognition.
\newblock In \emph{Proceedings of the IEEE/CVF Conference on Computer Vision and Pattern Recognition}, pp.\  18688--18698, 2024{\natexlab{b}}.

\bibitem[Chowdhery et~al.(2023)Chowdhery, Narang, Devlin, Bosma, Mishra, Roberts, Barham, Chung, Sutton, Gehrmann, et~al.]{chowdhery2023palm}
Aakanksha Chowdhery, Sharan Narang, Jacob Devlin, Maarten Bosma, Gaurav Mishra, Adam Roberts, Paul Barham, Hyung~Won Chung, Charles Sutton, Sebastian Gehrmann, et~al.
\newblock Palm: Scaling language modeling with pathways.
\newblock \emph{Journal of Machine Learning Research}, 24\penalty0 (240):\penalty0 1--113, 2023.

\bibitem[Deng et~al.(2009)Deng, Dong, Socher, Li, Li, and Fei-Fei]{deng2009imagenet}
Jia Deng, Wei Dong, Richard Socher, Li-Jia Li, Kai Li, and Li~Fei-Fei.
\newblock Imagenet: A large-scale hierarchical image database.
\newblock In \emph{2009 IEEE conference on computer vision and pattern recognition}, pp.\  248--255. Ieee, 2009.

\bibitem[Dosovitskiy et~al.(2020)Dosovitskiy, Beyer, Kolesnikov, Weissenborn, Zhai, Unterthiner, Dehghani, Minderer, Heigold, Gelly, et~al.]{dosovitskiy2020image}
Alexey Dosovitskiy, Lucas Beyer, Alexander Kolesnikov, Dirk Weissenborn, Xiaohua Zhai, Thomas Unterthiner, Mostafa Dehghani, Matthias Minderer, Georg Heigold, Sylvain Gelly, et~al.
\newblock An image is worth 16x16 words: Transformers for image recognition at scale.
\newblock \emph{arXiv preprint arXiv:2010.11929}, 2020.

\bibitem[Fan et~al.(2021)Fan, Xiong, Mangalam, Li, Yan, Malik, and Feichtenhofer]{fan2021multiscale}
Haoqi Fan, Bo~Xiong, Karttikeya Mangalam, Yanghao Li, Zhicheng Yan, Jitendra Malik, and Christoph Feichtenhofer.
\newblock Multiscale vision transformers.
\newblock In \emph{Proceedings of the IEEE/CVF international conference on computer vision}, pp.\  6824--6835, 2021.

\bibitem[Fan et~al.(2024)Fan, Krishnan, Isola, Katabi, and Tian]{fan2024improving}
Lijie Fan, Dilip Krishnan, Phillip Isola, Dina Katabi, and Yonglong Tian.
\newblock Improving clip training with language rewrites.
\newblock \emph{Advances in Neural Information Processing Systems}, 36, 2024.

\bibitem[Feichtenhofer(2020)]{feichtenhofer2020x3d}
Christoph Feichtenhofer.
\newblock X3d: Expanding architectures for efficient video recognition.
\newblock In \emph{Proceedings of the IEEE/CVF conference on computer vision and pattern recognition}, pp.\  203--213, 2020.

\bibitem[Feichtenhofer et~al.(2019)Feichtenhofer, Fan, Malik, and He]{feichtenhofer2019slowfast}
Christoph Feichtenhofer, Haoqi Fan, Jitendra Malik, and Kaiming He.
\newblock Slowfast networks for video recognition.
\newblock In \emph{Proceedings of the IEEE/CVF international conference on computer vision}, pp.\  6202--6211, 2019.

\bibitem[Feichtenhofer et~al.(2022)Feichtenhofer, Li, He, et~al.]{feichtenhofer2022masked}
Christoph Feichtenhofer, Yanghao Li, Kaiming He, et~al.
\newblock Masked autoencoders as spatiotemporal learners.
\newblock \emph{Advances in neural information processing systems}, 35:\penalty0 35946--35958, 2022.

\bibitem[Frome et~al.(2013)Frome, Corrado, Shlens, Bengio, Dean, Ranzato, and Mikolov]{frome2013devise}
Andrea Frome, Greg~S Corrado, Jon Shlens, Samy Bengio, Jeff Dean, Marc'Aurelio Ranzato, and Tomas Mikolov.
\newblock Devise: A deep visual-semantic embedding model.
\newblock \emph{Advances in neural information processing systems}, 26, 2013.

\bibitem[Gao et~al.(2019)Gao, Zhang, and Xu]{gao2019know}
Junyu Gao, Tianzhu Zhang, and Changsheng Xu.
\newblock I know the relationships: Zero-shot action recognition via two-stream graph convolutional networks and knowledge graphs.
\newblock In \emph{Proceedings of the AAAI conference on artificial intelligence}, volume~33, pp.\  8303--8311, 2019.

\bibitem[Ghosh et~al.(2020)Ghosh, Saini, Davis, and Shrivastava]{ghosh2020all}
Pallabi Ghosh, Nirat Saini, Larry~S Davis, and Abhinav Shrivastava.
\newblock All about knowledge graphs for actions.
\newblock \emph{arXiv preprint arXiv:2008.12432}, 2020.

\bibitem[Girdhar \& Grauman(2021)Girdhar and Grauman]{girdhar2021anticipative}
Rohit Girdhar and Kristen Grauman.
\newblock Anticipative video transformer.
\newblock In \emph{Proceedings of the IEEE/CVF international conference on computer vision}, pp.\  13505--13515, 2021.

\bibitem[Gu et~al.(2021)Gu, Lin, Kuo, and Cui]{gu2021open}
Xiuye Gu, Tsung-Yi Lin, Weicheng Kuo, and Yin Cui.
\newblock Open-vocabulary object detection via vision and language knowledge distillation.
\newblock \emph{arXiv preprint arXiv:2104.13921}, 2021.

\bibitem[Guo et~al.(2021)Guo, Guo, and Lu]{guo2021ssan}
Xudong Guo, Xun Guo, and Yan Lu.
\newblock Ssan: Separable self-attention network for video representation learning.
\newblock In \emph{Proceedings of the IEEE/CVF conference on computer vision and pattern recognition}, pp.\  12618--12627, 2021.

\bibitem[Han et~al.(2023)Han, Pan, Han, Song, and Huang]{han2023flatten}
Dongchen Han, Xuran Pan, Yizeng Han, Shiji Song, and Gao Huang.
\newblock Flatten transformer: Vision transformer using focused linear attention.
\newblock In \emph{Proceedings of the IEEE/CVF International Conference on Computer Vision}, pp.\  5961--5971, 2023.

\bibitem[Hara et~al.(2017)Hara, Kataoka, and Satoh]{hara2017learning}
Kensho Hara, Hirokatsu Kataoka, and Yutaka Satoh.
\newblock Learning spatio-temporal features with 3d residual networks for action recognition.
\newblock In \emph{Proceedings of the IEEE international conference on computer vision workshops}, pp.\  3154--3160, 2017.

\bibitem[Hendricks \& Nematzadeh(2021)Hendricks and Nematzadeh]{hendricks2021probing}
Lisa~Anne Hendricks and Aida Nematzadeh.
\newblock Probing image-language transformers for verb understanding.
\newblock \emph{arXiv preprint arXiv:2106.09141}, 2021.

\bibitem[Jia et~al.(2021)Jia, Yang, Xia, Chen, Parekh, Pham, Le, Sung, Li, and Duerig]{jia2021scaling}
Chao Jia, Yinfei Yang, Ye~Xia, Yi-Ting Chen, Zarana Parekh, Hieu Pham, Quoc Le, Yun-Hsuan Sung, Zhen Li, and Tom Duerig.
\newblock Scaling up visual and vision-language representation learning with noisy text supervision.
\newblock In \emph{International conference on machine learning}, pp.\  4904--4916. PMLR, 2021.

\bibitem[Ju et~al.(2022)Ju, Han, Zheng, Zhang, and Xie]{ju2022prompting}
Chen Ju, Tengda Han, Kunhao Zheng, Ya~Zhang, and Weidi Xie.
\newblock Prompting visual-language models for efficient video understanding.
\newblock In \emph{European Conference on Computer Vision}, pp.\  105--124. Springer, 2022.

\bibitem[Karpathy et~al.(2014)Karpathy, Toderici, Shetty, Leung, Sukthankar, and Fei-Fei]{karpathy2014large}
Andrej Karpathy, George Toderici, Sanketh Shetty, Thomas Leung, Rahul Sukthankar, and Li~Fei-Fei.
\newblock Large-scale video classification with convolutional neural networks.
\newblock In \emph{Proceedings of the IEEE conference on Computer Vision and Pattern Recognition}, pp.\  1725--1732, 2014.

\bibitem[Kay et~al.(2017)Kay, Carreira, Simonyan, Zhang, Hillier, Vijayanarasimhan, Viola, Green, Back, Natsev, et~al.]{kay2017kinetics}
Will Kay, Joao Carreira, Karen Simonyan, Brian Zhang, Chloe Hillier, Sudheendra Vijayanarasimhan, Fabio Viola, Tim Green, Trevor Back, Paul Natsev, et~al.
\newblock The kinetics human action video dataset.
\newblock \emph{arXiv preprint arXiv:1705.06950}, 2017.

\bibitem[Kuehne et~al.(2011)Kuehne, Jhuang, Garrote, Poggio, and Serre]{kuehne2011hmdb}
Hildegard Kuehne, Hueihan Jhuang, Est{\'\i}baliz Garrote, Tomaso Poggio, and Thomas Serre.
\newblock Hmdb: a large video database for human motion recognition.
\newblock In \emph{2011 International conference on computer vision}, pp.\  2556--2563. IEEE, 2011.

\bibitem[Li et~al.(2020{\natexlab{a}})Li, Duan, Fang, Gong, and Jiang]{li2020unicoder}
Gen Li, Nan Duan, Yuejian Fang, Ming Gong, and Daxin Jiang.
\newblock Unicoder-vl: A universal encoder for vision and language by cross-modal pre-training.
\newblock In \emph{Proceedings of the AAAI conference on artificial intelligence}, volume~34, pp.\  11336--11344, 2020{\natexlab{a}}.

\bibitem[Li et~al.(2022{\natexlab{a}})Li, Wang, Gao, Song, Liu, Li, and Qiao]{li2022uniformer}
Kunchang Li, Yali Wang, Peng Gao, Guanglu Song, Yu~Liu, Hongsheng Li, and Yu~Qiao.
\newblock Uniformer: Unified transformer for efficient spatiotemporal representation learning.
\newblock \emph{arXiv preprint arXiv:2201.04676}, 2022{\natexlab{a}}.

\bibitem[Li et~al.(2022{\natexlab{b}})Li, Zhang, Zhang, Yang, Li, Zhong, Wang, Yuan, Zhang, Hwang, et~al.]{li2022grounded}
Liunian~Harold Li, Pengchuan Zhang, Haotian Zhang, Jianwei Yang, Chunyuan Li, Yiwu Zhong, Lijuan Wang, Lu~Yuan, Lei Zhang, Jenq-Neng Hwang, et~al.
\newblock Grounded language-image pre-training.
\newblock In \emph{Proceedings of the IEEE/CVF Conference on Computer Vision and Pattern Recognition}, pp.\  10965--10975, 2022{\natexlab{b}}.

\bibitem[Li et~al.()Li, Wu, Fan, Mangalam, Xiong, Malik, and Feichtenhofer]{li2112improved}
Y~Li, CY~Wu, H~Fan, K~Mangalam, B~Xiong, J~Malik, and C~Feichtenhofer.
\newblock Improved multiscale vision transformers for classification and detection. arxiv 2021.
\newblock \emph{arXiv preprint arXiv:2112.01526}.

\bibitem[Li et~al.(2020{\natexlab{b}})Li, Ji, Shi, Zhang, Kang, and Wang]{li2020tea}
Yan Li, Bin Ji, Xintian Shi, Jianguo Zhang, Bin Kang, and Limin Wang.
\newblock Tea: Temporal excitation and aggregation for action recognition.
\newblock In \emph{Proceedings of the IEEE/CVF conference on computer vision and pattern recognition}, pp.\  909--918, 2020{\natexlab{b}}.

\bibitem[Lin et~al.(2019)Lin, Gan, and Han]{lin2019tsm}
Ji~Lin, Chuang Gan, and Song Han.
\newblock Tsm: Temporal shift module for efficient video understanding.
\newblock In \emph{Proceedings of the IEEE/CVF international conference on computer vision}, pp.\  7083--7093, 2019.

\bibitem[Lin et~al.(2023)Lin, Karlinsky, Shvetsova, Possegger, Kozinski, Panda, Feris, Kuehne, and Bischof]{lin2023match}
Wei Lin, Leonid Karlinsky, Nina Shvetsova, Horst Possegger, Mateusz Kozinski, Rameswar Panda, Rogerio Feris, Hilde Kuehne, and Horst Bischof.
\newblock Match, expand and improve: Unsupervised finetuning for zero-shot action recognition with language knowledge.
\newblock In \emph{Proceedings of the IEEE/CVF International Conference on Computer Vision}, pp.\  2851--2862, 2023.

\bibitem[Lin et~al.(2022)Lin, Geng, Zhang, Gao, de~Melo, Wang, Dai, Qiao, and Li]{lin2022frozen}
Ziyi Lin, Shijie Geng, Renrui Zhang, Peng Gao, Gerard de~Melo, Xiaogang Wang, Jifeng Dai, Yu~Qiao, and Hongsheng Li.
\newblock Frozen clip models are efficient video learners.
\newblock In \emph{European Conference on Computer Vision}, pp.\  388--404. Springer, 2022.

\bibitem[Liu et~al.(2022)Liu, Ning, Cao, Wei, Zhang, Lin, and Hu]{liu2022video}
Ze~Liu, Jia Ning, Yue Cao, Yixuan Wei, Zheng Zhang, Stephen Lin, and Han Hu.
\newblock Video swin transformer.
\newblock In \emph{Proceedings of the IEEE/CVF conference on computer vision and pattern recognition}, pp.\  3202--3211, 2022.

\bibitem[Liu et~al.(2021)Liu, Wang, Wu, Qian, and Lu]{liu2021tam}
Zhaoyang Liu, Limin Wang, Wayne Wu, Chen Qian, and Tong Lu.
\newblock Tam: Temporal adaptive module for video recognition.
\newblock In \emph{Proceedings of the IEEE/CVF international conference on computer vision}, pp.\  13708--13718, 2021.

\bibitem[Lu et~al.(2019)Lu, Batra, Parikh, and Lee]{lu2019vilbert}
Jiasen Lu, Dhruv Batra, Devi Parikh, and Stefan Lee.
\newblock Vilbert: Pretraining task-agnostic visiolinguistic representations for vision-and-language tasks.
\newblock \emph{Advances in neural information processing systems}, 32, 2019.

\bibitem[Luo et~al.(2022)Luo, Ji, Zhong, Chen, Lei, Duan, and Li]{luo2022clip4clip}
Huaishao Luo, Lei Ji, Ming Zhong, Yang Chen, Wen Lei, Nan Duan, and Tianrui Li.
\newblock Clip4clip: An empirical study of clip for end to end video clip retrieval and captioning.
\newblock \emph{Neurocomputing}, 508:\penalty0 293--304, 2022.

\bibitem[Miech et~al.(2019)Miech, Zhukov, Alayrac, Tapaswi, Laptev, and Sivic]{miech2019howto100m}
Antoine Miech, Dimitri Zhukov, Jean-Baptiste Alayrac, Makarand Tapaswi, Ivan Laptev, and Josef Sivic.
\newblock Howto100m: Learning a text-video embedding by watching hundred million narrated video clips.
\newblock In \emph{Proceedings of the IEEE/CVF international conference on computer vision}, pp.\  2630--2640, 2019.

\bibitem[Momeni et~al.(2023)Momeni, Caron, Nagrani, Zisserman, and Schmid]{momeni2023verbs}
Liliane Momeni, Mathilde Caron, Arsha Nagrani, Andrew Zisserman, and Cordelia Schmid.
\newblock Verbs in action: Improving verb understanding in video-language models.
\newblock In \emph{Proceedings of the IEEE/CVF International Conference on Computer Vision}, pp.\  15579--15591, 2023.

\bibitem[Neimark et~al.(2021)Neimark, Bar, Zohar, and Asselmann]{neimark2021video}
Daniel Neimark, Omri Bar, Maya Zohar, and Dotan Asselmann.
\newblock Video transformer network.
\newblock In \emph{Proceedings of the IEEE/CVF international conference on computer vision}, pp.\  3163--3172, 2021.

\bibitem[Ni et~al.(2022)Ni, Peng, Chen, Zhang, Meng, Fu, Xiang, and Ling]{ni2022expanding}
Bolin Ni, Houwen Peng, Minghao Chen, Songyang Zhang, Gaofeng Meng, Jianlong Fu, Shiming Xiang, and Haibin Ling.
\newblock Expanding language-image pretrained models for general video recognition.
\newblock In \emph{European Conference on Computer Vision}, pp.\  1--18. Springer, 2022.

\bibitem[OpenAI(2024)]{openai2024gpt4technicalreport}
OpenAI.
\newblock Gpt-4 technical report, https://arxiv.org/abs/2303.08774, 2024.
\newblock URL \url{https://chat.openai.com}.

\bibitem[Pan et~al.(2022)Pan, Lin, Zhu, Shao, and Li]{pan2022st}
Junting Pan, Ziyi Lin, Xiatian Zhu, Jing Shao, and Hongsheng Li.
\newblock St-adapter: Parameter-efficient image-to-video transfer learning.
\newblock \emph{Advances in Neural Information Processing Systems}, 35:\penalty0 26462--26477, 2022.

\bibitem[Patrick et~al.(2021)Patrick, Campbell, Asano, Misra, Metze, Feichtenhofer, Vedaldi, and Henriques]{patrick2021keeping}
Mandela Patrick, Dylan Campbell, Yuki Asano, Ishan Misra, Florian Metze, Christoph Feichtenhofer, Andrea Vedaldi, and Joao~F Henriques.
\newblock Keeping your eye on the ball: Trajectory attention in video transformers.
\newblock \emph{Advances in neural information processing systems}, 34:\penalty0 12493--12506, 2021.

\bibitem[Qian et~al.(2022)Qian, Yu, Liu, and Hauptmann]{qian2022rethinking}
Yijun Qian, Lijun Yu, Wenhe Liu, and Alexander~G Hauptmann.
\newblock Rethinking zero-shot action recognition: Learning from latent atomic actions.
\newblock In \emph{European Conference on Computer Vision}, pp.\  104--120. Springer, 2022.

\bibitem[Qin et~al.(2017)Qin, Liu, Shao, Shen, Ni, Chen, and Wang]{qin2017zero}
Jie Qin, Li~Liu, Ling Shao, Fumin Shen, Bingbing Ni, Jiaxin Chen, and Yunhong Wang.
\newblock Zero-shot action recognition with error-correcting output codes.
\newblock In \emph{Proceedings of the IEEE Conference on Computer Vision and Pattern Recognition}, pp.\  2833--2842, 2017.

\bibitem[Qiu et~al.(2017)Qiu, Yao, and Mei]{qiu2017learning}
Zhaofan Qiu, Ting Yao, and Tao Mei.
\newblock Learning spatio-temporal representation with pseudo-3d residual networks.
\newblock In \emph{proceedings of the IEEE International Conference on Computer Vision}, pp.\  5533--5541, 2017.

\bibitem[Radford et~al.(2021)Radford, Kim, Hallacy, Ramesh, Goh, Agarwal, Sastry, Askell, Mishkin, Clark, et~al.]{radford2021learning}
Alec Radford, Jong~Wook Kim, Chris Hallacy, Aditya Ramesh, Gabriel Goh, Sandhini Agarwal, Girish Sastry, Amanda Askell, Pamela Mishkin, Jack Clark, et~al.
\newblock Learning transferable visual models from natural language supervision.
\newblock In \emph{International conference on machine learning}, pp.\  8748--8763. PMLR, 2021.

\bibitem[Rasheed et~al.(2023)Rasheed, Khattak, Maaz, Khan, and Khan]{rasheed2023fine}
Hanoona Rasheed, Muhammad~Uzair Khattak, Muhammad Maaz, Salman Khan, and Fahad~Shahbaz Khan.
\newblock Fine-tuned clip models are efficient video learners.
\newblock In \emph{Proceedings of the IEEE/CVF Conference on Computer Vision and Pattern Recognition}, pp.\  6545--6554, 2023.

\bibitem[Romera-Paredes \& Torr(2015)Romera-Paredes and Torr]{romera2015embarrassingly}
Bernardino Romera-Paredes and Philip Torr.
\newblock An embarrassingly simple approach to zero-shot learning.
\newblock In \emph{International conference on machine learning}, pp.\  2152--2161. PMLR, 2015.

\bibitem[Ryoo et~al.(2021)Ryoo, Piergiovanni, Arnab, Dehghani, and Angelova]{ryoo2021tokenlearner}
Michael Ryoo, AJ~Piergiovanni, Anurag Arnab, Mostafa Dehghani, and Anelia Angelova.
\newblock Tokenlearner: Adaptive space-time tokenization for videos.
\newblock \emph{Advances in Neural Information Processing Systems}, 34:\penalty0 12786--12797, 2021.

\bibitem[Saharia et~al.(2022)Saharia, Chan, Saxena, Li, Whang, Denton, Ghasemipour, Gontijo~Lopes, Karagol~Ayan, Salimans, et~al.]{saharia2022photorealistic}
Chitwan Saharia, William Chan, Saurabh Saxena, Lala Li, Jay Whang, Emily~L Denton, Kamyar Ghasemipour, Raphael Gontijo~Lopes, Burcu Karagol~Ayan, Tim Salimans, et~al.
\newblock Photorealistic text-to-image diffusion models with deep language understanding.
\newblock \emph{Advances in Neural Information Processing Systems}, 35:\penalty0 36479--36494, 2022.

\bibitem[Simonyan \& Zisserman(2014)Simonyan and Zisserman]{simonyan2014two}
Karen Simonyan and Andrew Zisserman.
\newblock Two-stream convolutional networks for action recognition in videos.
\newblock \emph{Advances in neural information processing systems}, 27, 2014.

\bibitem[Soomro et~al.(2012)Soomro, Zamir, and Shah]{soomro2012ucf101}
Khurram Soomro, Amir~Roshan Zamir, and Mubarak Shah.
\newblock Ucf101: A dataset of 101 human actions classes from videos in the wild.
\newblock \emph{arXiv preprint arXiv:1212.0402}, 2012.

\bibitem[Su et~al.(2019)Su, Zhu, Cao, Li, Lu, Wei, and Dai]{su2019vl}
Weijie Su, Xizhou Zhu, Yue Cao, Bin Li, Lewei Lu, Furu Wei, and Jifeng Dai.
\newblock Vl-bert: Pre-training of generic visual-linguistic representations.
\newblock \emph{arXiv preprint arXiv:1908.08530}, 2019.

\bibitem[Sun et~al.(2019{\natexlab{a}})Sun, Baradel, Murphy, and Schmid]{sun2019learning}
Chen Sun, Fabien Baradel, Kevin Murphy, and Cordelia Schmid.
\newblock Learning video representations using contrastive bidirectional transformer.
\newblock \emph{arXiv preprint arXiv:1906.05743}, 2019{\natexlab{a}}.

\bibitem[Sun et~al.(2019{\natexlab{b}})Sun, Myers, Vondrick, Murphy, and Schmid]{sun2019videobert}
Chen Sun, Austin Myers, Carl Vondrick, Kevin Murphy, and Cordelia Schmid.
\newblock Videobert: A joint model for video and language representation learning.
\newblock In \emph{Proceedings of the IEEE/CVF international conference on computer vision}, pp.\  7464--7473, 2019{\natexlab{b}}.

\bibitem[Tan \& Bansal(2019)Tan and Bansal]{tan2019lxmert}
Hao Tan and Mohit Bansal.
\newblock Lxmert: Learning cross-modality encoder representations from transformers.
\newblock \emph{arXiv preprint arXiv:1908.07490}, 2019.

\bibitem[Thrush et~al.(2022)Thrush, Jiang, Bartolo, Singh, Williams, Kiela, and Ross]{thrush2022winoground}
Tristan Thrush, Ryan Jiang, Max Bartolo, Amanpreet Singh, Adina Williams, Douwe Kiela, and Candace Ross.
\newblock Winoground: Probing vision and language models for visio-linguistic compositionality.
\newblock In \emph{Proceedings of the IEEE/CVF Conference on Computer Vision and Pattern Recognition}, pp.\  5238--5248, 2022.

\bibitem[Tong et~al.(2022)Tong, Song, Wang, and Wang]{tong2022videomae}
Zhan Tong, Yibing Song, Jue Wang, and Limin Wang.
\newblock Videomae: Masked autoencoders are data-efficient learners for self-supervised video pre-training.
\newblock \emph{Advances in neural information processing systems}, 35:\penalty0 10078--10093, 2022.

\bibitem[Touvron et~al.(2023{\natexlab{a}})Touvron, Lavril, Izacard, Martinet, Lachaux, Lacroix, Rozi{\`e}re, Goyal, Hambro, Azhar, et~al.]{touvron2023llama}
Hugo Touvron, Thibaut Lavril, Gautier Izacard, Xavier Martinet, Marie-Anne Lachaux, Timoth{\'e}e Lacroix, Baptiste Rozi{\`e}re, Naman Goyal, Eric Hambro, Faisal Azhar, et~al.
\newblock Llama: Open and efficient foundation language models.
\newblock \emph{arXiv preprint arXiv:2302.13971}, 2023{\natexlab{a}}.

\bibitem[Touvron et~al.(2023{\natexlab{b}})Touvron, Martin, Stone, Albert, Almahairi, Babaei, Bashlykov, Batra, Bhargava, Bhosale, et~al.]{touvron2023llama2}
Hugo Touvron, Louis Martin, Kevin Stone, Peter Albert, Amjad Almahairi, Yasmine Babaei, Nikolay Bashlykov, Soumya Batra, Prajjwal Bhargava, Shruti Bhosale, et~al.
\newblock Llama 2: Open foundation and fine-tuned chat models.
\newblock \emph{arXiv preprint arXiv:2307.09288}, 2023{\natexlab{b}}.

\bibitem[Tran et~al.(2015)Tran, Bourdev, Fergus, Torresani, and Paluri]{tran2015learning}
Du~Tran, Lubomir Bourdev, Rob Fergus, Lorenzo Torresani, and Manohar Paluri.
\newblock Learning spatiotemporal features with 3d convolutional networks.
\newblock In \emph{Proceedings of the IEEE international conference on computer vision}, pp.\  4489--4497, 2015.

\bibitem[Tran et~al.(2018)Tran, Wang, Torresani, Ray, LeCun, and Paluri]{tran2018closer}
Du~Tran, Heng Wang, Lorenzo Torresani, Jamie Ray, Yann LeCun, and Manohar Paluri.
\newblock A closer look at spatiotemporal convolutions for action recognition.
\newblock In \emph{Proceedings of the IEEE conference on Computer Vision and Pattern Recognition}, pp.\  6450--6459, 2018.

\bibitem[Tu et~al.(2023)Tu, Dai, Wu, Cheng, Hu, and Jiang]{tu2023implicit}
Shuyuan Tu, Qi~Dai, Zuxuan Wu, Zhi-Qi Cheng, Han Hu, and Yu-Gang Jiang.
\newblock Implicit temporal modeling with learnable alignment for video recognition.
\newblock In \emph{Proceedings of the IEEE/CVF International Conference on Computer Vision}, pp.\  19936--19947, 2023.

\bibitem[Vinker et~al.(2022)Vinker, Pajouheshgar, Bo, Bachmann, Bermano, Cohen-Or, Zamir, and Shamir]{vinker2022clipasso}
Yael Vinker, Ehsan Pajouheshgar, Jessica~Y Bo, Roman~Christian Bachmann, Amit~Haim Bermano, Daniel Cohen-Or, Amir Zamir, and Ariel Shamir.
\newblock Clipasso: Semantically-aware object sketching.
\newblock \emph{ACM Transactions on Graphics (TOG)}, 41\penalty0 (4):\penalty0 1--11, 2022.

\bibitem[Wang et~al.(2016)Wang, Xiong, Wang, Qiao, Lin, Tang, and Van~Gool]{wang2016temporal}
Limin Wang, Yuanjun Xiong, Zhe Wang, Yu~Qiao, Dahua Lin, Xiaoou Tang, and Luc Van~Gool.
\newblock Temporal segment networks: Towards good practices for deep action recognition.
\newblock In \emph{European conference on computer vision}, pp.\  20--36. Springer, 2016.

\bibitem[Wang et~al.(2021)Wang, Xing, and Liu]{wang2021actionclip}
Mengmeng Wang, Jiazheng Xing, and Yong Liu.
\newblock Actionclip: A new paradigm for video action recognition.
\newblock \emph{arXiv preprint arXiv:2109.08472}, 2021.

\bibitem[Wang \& Chen(2017)Wang and Chen]{wang2017alternative}
Qian Wang and Ke~Chen.
\newblock Alternative semantic representations for zero-shot human action recognition.
\newblock In \emph{Machine Learning and Knowledge Discovery in Databases: European Conference, ECML PKDD 2017, Skopje, Macedonia, September 18--22, 2017, Proceedings, Part I 10}, pp.\  87--102. Springer, 2017.

\bibitem[Wang et~al.(2018)Wang, Girshick, Gupta, and He]{wang2018non}
Xiaolong Wang, Ross Girshick, Abhinav Gupta, and Kaiming He.
\newblock Non-local neural networks.
\newblock In \emph{Proceedings of the IEEE conference on computer vision and pattern recognition}, pp.\  7794--7803, 2018.

\bibitem[Xie et~al.(2018)Xie, Sun, Huang, Tu, and Murphy]{xie2018rethinking}
Saining Xie, Chen Sun, Jonathan Huang, Zhuowen Tu, and Kevin Murphy.
\newblock Rethinking spatiotemporal feature learning: Speed-accuracy trade-offs in video classification.
\newblock In \emph{Proceedings of the European conference on computer vision (ECCV)}, pp.\  305--321, 2018.

\bibitem[Xie et~al.(2024)Xie, Cui, Tan, Zheng, and Yu]{xie2024fusionmamba}
Xinyu Xie, Yawen Cui, Tao Tan, Xubin Zheng, and Zitong Yu.
\newblock Fusionmamba: Dynamic feature enhancement for multimodal image fusion with mamba.
\newblock \emph{Visual Intelligence}, 2\penalty0 (37), 2024.

\bibitem[Xu et~al.(2021)Xu, Ghosh, Huang, Okhonko, Aghajanyan, Metze, Zettlemoyer, and Feichtenhofer]{xu2021videoclip}
Hu~Xu, Gargi Ghosh, Po-Yao Huang, Dmytro Okhonko, Armen Aghajanyan, Florian Metze, Luke Zettlemoyer, and Christoph Feichtenhofer.
\newblock Videoclip: Contrastive pre-training for zero-shot video-text understanding.
\newblock \emph{arXiv preprint arXiv:2109.14084}, 2021.

\bibitem[Xu et~al.(2022)Xu, De~Mello, Liu, Byeon, Breuel, Kautz, and Wang]{xu2022groupvit}
Jiarui Xu, Shalini De~Mello, Sifei Liu, Wonmin Byeon, Thomas Breuel, Jan Kautz, and Xiaolong Wang.
\newblock Groupvit: Semantic segmentation emerges from text supervision.
\newblock In \emph{Proceedings of the IEEE/CVF Conference on Computer Vision and Pattern Recognition}, pp.\  18134--18144, 2022.

\bibitem[Xu et~al.(2016)Xu, Hospedales, and Gong]{xu2016multi}
Xun Xu, Timothy~M Hospedales, and Shaogang Gong.
\newblock Multi-task zero-shot action recognition with prioritised data augmentation.
\newblock In \emph{Computer Vision--ECCV 2016: 14th European Conference, Amsterdam, The Netherlands, October 11-14, 2016, Proceedings, Part II 14}, pp.\  343--359. Springer, 2016.

\bibitem[Xue et~al.(2022)Xue, Sun, Liu, Fu, Song, Li, and Luo]{xue2022clip}
Hongwei Xue, Yuchong Sun, Bei Liu, Jianlong Fu, Ruihua Song, Houqiang Li, and Jiebo Luo.
\newblock Clip-vip: Adapting pre-trained image-text model to video-language alignment.
\newblock In \emph{The Eleventh International Conference on Learning Representations}, 2022.

\bibitem[Yan et~al.(2022)Yan, Xiong, Arnab, Lu, Zhang, Sun, and Schmid]{yan2022multiview}
Shen Yan, Xuehan Xiong, Anurag Arnab, Zhichao Lu, Mi~Zhang, Chen Sun, and Cordelia Schmid.
\newblock Multiview transformers for video recognition.
\newblock In \emph{Proceedings of the IEEE/CVF conference on computer vision and pattern recognition}, pp.\  3333--3343, 2022.

\bibitem[Yang et~al.(2023)Yang, Zhu, Xie, Zhang, Chen, and Li]{yang2023aim}
Taojiannan Yang, Yi~Zhu, Yusheng Xie, Aston Zhang, Chen Chen, and Mu~Li.
\newblock Aim: Adapting image models for efficient video action recognition.
\newblock \emph{arXiv preprint arXiv:2302.03024}, 2023.

\bibitem[Yu et~al.(2021)Yu, Zhou, Wan, Wang, Chen, Liu, Li, and Zhao]{yu2021searching}
Zitong Yu, Benjia Zhou, Jun Wan, Pichao Wang, Haoyu Chen, Xin Liu, Stan~Z Li, and Guoying Zhao.
\newblock Searching multi-rate and multi-modal temporal enhanced networks for gesture recognition.
\newblock \emph{IEEE Transactions on Image Processing}, 30:\penalty0 5626--5640, 2021.

\bibitem[Yuan et~al.(2021)Yuan, Chen, Chen, Codella, Dai, Gao, Hu, Huang, Li, Li, et~al.]{yuan2021florence}
Lu~Yuan, Dongdong Chen, Yi-Ling Chen, Noel Codella, Xiyang Dai, Jianfeng Gao, Houdong Hu, Xuedong Huang, Boxin Li, Chunyuan Li, et~al.
\newblock Florence: A new foundation model for computer vision.
\newblock \emph{arXiv preprint arXiv:2111.11432}, 2021.

\bibitem[Zhang et~al.(2021)Zhang, Yu, Fifty, Han, Dai, Pang, and Sha]{zhang2021co}
Bowen Zhang, Jiahui Yu, Christopher Fifty, Wei Han, Andrew~M Dai, Ruoming Pang, and Fei Sha.
\newblock Co-training transformer with videos and images improves action recognition.
\newblock \emph{arXiv preprint arXiv:2112.07175}, 2021.

\bibitem[Zhang et~al.(2017)Zhang, Xiang, and Gong]{zhang2017learning}
Li~Zhang, Tao Xiang, and Shaogang Gong.
\newblock Learning a deep embedding model for zero-shot learning.
\newblock In \emph{Proceedings of the IEEE conference on computer vision and pattern recognition}, pp.\  2021--2030, 2017.

\bibitem[Zhao et~al.(2022)Zhao, Li, Guo, and Lu]{zhao2022alignment}
Yizhou Zhao, Zhenyang Li, Xun Guo, and Yan Lu.
\newblock Alignment-guided temporal attention for video action recognition.
\newblock \emph{Advances in Neural Information Processing Systems}, 35:\penalty0 13627--13639, 2022.

\bibitem[Zhou et~al.(2018)Zhou, Andonian, Oliva, and Torralba]{zhou2018temporal}
Bolei Zhou, Alex Andonian, Aude Oliva, and Antonio Torralba.
\newblock Temporal relational reasoning in videos.
\newblock In \emph{Proceedings of the European conference on computer vision (ECCV)}, pp.\  803--818, 2018.

\bibitem[Zhu \& Yang(2020)Zhu and Yang]{zhu2020actbert}
Linchao Zhu and Yi~Yang.
\newblock Actbert: Learning global-local video-text representations.
\newblock In \emph{Proceedings of the IEEE/CVF conference on computer vision and pattern recognition}, pp.\  8746--8755, 2020.

\bibitem[Zhu et~al.(2018)Zhu, Long, Guan, Newsam, and Shao]{zhu2018towards}
Yi~Zhu, Yang Long, Yu~Guan, Shawn Newsam, and Ling Shao.
\newblock Towards universal representation for unseen action recognition.
\newblock In \emph{Proceedings of the IEEE conference on computer vision and pattern recognition}, pp.\  9436--9445, 2018.

\end{thebibliography}
\bibliographystyle{iclr2025_conference}

\newpage
\section{Appendix}
\appendix
\textbf{Roadmap}

\label{sec: roadmap}
In the Appendix, we present proof in \ref{sec: Proof}, Spatiotemporal homogenization study details in \ref{sec: Spatiotemporal Homogenization Study}, some additional experiment results in \ref{sec: Supplementary Experiments}, experiment setting details in \ref{sec: Experiment Setting Details}, interpretive prompt technique details in \ref{sec: Interpretive Prompt Details}, analysis of synthetic video testing in \ref{sec:Analysis of Synthetic Video Testing}, dataset details in \ref{sec: Dataset Details}, and more visualization of spatiotemporal attention map in \ref{sec: More Attention Heat Map Visualization}.


\section{Proof}
\label{sec: Proof}
We conclude that there is a low-rank bottleneck problem for joint attention and KMTA when facing longer token sequence lengths (\(d \ll n\)), while KMCTA always guarantees the property of full rank under our assumption. Proofs as follow:

\subsection{Proof of joint attention cannot guarantee full rank}

Based on the Representation Theorem of Bhojanapalli et al. \cite{bhojanapalli2020low}:

\textbf{Theorem 1} (Representation Theorem \cite{bhojanapalli2020low}). \textit{If} \(d_q=d_k=d \geq n\), \textit{then given any full row rank matrix}  \(\textbf{X} \in \mathcal{R}^{n \times d}\) \textit{and an arbitrary} \(n \times n\) \textit{positive row stochastic matrix} \(\textbf{P}\), \textit{subject to the sum of each row of} \(\textbf{P}\) \textit{is equal to 1, i.e.,} \(\textbf{P}\textbf{1}=\textbf{1}\), \textit{there always exists \(d \times d\) projection matrices} \(\textbf{W}_q\) \textit{and} \(\textbf{W}_k\) \textit{such that}
\begin{equation}
\label{eqn: representation theorem}
\mathrm{Softmax}[\frac{(\textbf{X}\textbf{W}_q)(\textbf{X}\textbf{W}_k)^T}{\sqrt{d_k}}] = \textbf{P}
\end{equation}
\textit{If} \(d_q=d_k=d < n\), \textit{there exist} \(\textbf{X}\) \textit{and} \(\textbf{P}\) \textit{such that Eqn. \ref{eqn: representation theorem} does not hold for all}  \(\textbf{W}_q\) \textit{and} \(\textbf{W}_k\).

The proof process of \textbf{Theorem 1} are referred from Bhojanapalli et al. \cite{bhojanapalli2020low}. By the way, in Bhojanapalli et al. \cite{bhojanapalli2020low}, they set \(\textbf{X} \in \mathcal{R}^{d \times n}\). However, for the convenience of our subsequent proof, we transpose the \(n\) and \(d\) dimensions, \(\textbf{X} \in \mathcal{R}^{n \times d}\). 

For \(d \geq n\) case. Since \(\textbf{X}\) exhibits full row rank, there exists a right pseudo inverse \(\textbf{X}^\dagger =\textbf{X}^T(\textbf{X}\textbf{X}^T)^{-1} \in \mathcal{R}^{d \times n}\) such that \(\textbf{X}\textbf{X}^\dagger=\textbf{I}_n\). Let \(\textbf{W}_k=\textbf{X}^\dagger\tilde{\textbf{W}}_k\) and \(\textbf{W}_q=\textbf{X}^\dagger\tilde{\textbf{W}}_q\). Then
\begin{align}
\label{eqn: representation theorem proof 1}
(\textbf{X}\textbf{W}_q)(\textbf{X}\textbf{W}_k)^T &= \textbf{X}\textbf{W}_q\textbf{W}_k^T\textbf{X}^T \\
& = \textbf{X}\textbf{X}^\dagger\tilde{\textbf{W}}_q\tilde{\textbf{W}}_k^T(\textbf{X}^\dagger)^T\textbf{X}^T \\
& = \textbf{X}\textbf{X}^\dagger\tilde{\textbf{W}}_q\tilde{\textbf{W}}_k^T(\textbf{X}\textbf{X}^\dagger)^T \\
& = \textbf{I}_n\tilde{\textbf{W}}_q\tilde{\textbf{W}}_k^T\textbf{I}_n^T \\
& = \tilde{\textbf{W}}_q\tilde{\textbf{W}}_k^T = \tilde{\textbf{W}}_{qk}
\end{align} 
According Eqn. \ref{eqn: representation theorem}, we obtain that
\begin{align}
\label{eqn: representation theorem proof 2}
\mathrm{Softmax}[\frac{(\textbf{X}\textbf{W}_q)(\textbf{X}\textbf{W}_k)^T}{\sqrt{d_k}}] & = \mathrm{Softmax}[\frac{\tilde{\textbf{W}}_{qk}}{\sqrt{d_k}}] \\
& =\textbf{D}^{-1}_{\tilde{\textbf{W}}_{qk}} \mathrm{exp}(\frac{\tilde{\textbf{W}}_{qk}}{\sqrt{d_k}})
\end{align} 
where \(\textbf{D}_{\tilde{\textbf{W}}_{qk}}\) is a \(n \times n\) diagonal matrix such that
\begin{align}
\label{eqn: representation theorem proof 3}
(\textbf{D}_{\tilde{\textbf{W}}_{qk}})_{ii} & =  \sum^n_{j=1} \mathrm{exp}(\frac{(\tilde{\textbf{W}}_{qk})_{ji}}{\sqrt{d_k}}) \\
& = (\mathrm{exp}(\frac{\tilde{\textbf{W}}_{qk}}{\sqrt{d_k}})\textbf{1})_{i}
\end{align} 
We can establish the desired result by showing that there always exists a \(\tilde{\textbf{W}}_{qk}\) that satisfies the following fixed point equation:
\begin{equation}
\label{eqn: representation theorem proof 4}
\textbf{D}_{\tilde{\textbf{W}}_{qk}}^{-1} \mathrm{exp}(\frac{\tilde{\textbf{W}}_{qk}}{\sqrt{d_k}}) = \textbf{P} 
\end{equation} 

Given \(\textbf{P}\), to construct such a \(\tilde{\textbf{W}}_{qk}\), we can pick an random positive diagonal matrix \(\textbf{D}_0\):
\begin{equation}
\label{eqn: representation theorem proof 5}
\tilde{\textbf{W}}_{qk} = \sqrt{d_k} \cdot \mathrm{log}(\textbf{D}_0\textbf{P})
\end{equation} 
Since \(\textbf{P}\) is a positive matrix, and \(\textbf{D}_0\) is a positive diagonal matrix, such a \(\tilde{\textbf{W}}_{qk}\) always exists. According to Eqn. \ref{eqn: representation theorem proof 3} we can conclude
\begin{align}
\label{eqn: representation theorem proof 6}
\textbf{D}_{\tilde{\textbf{W}}_{qk}} & = \mathrm{Diag}(\mathrm{exp}(\frac{\tilde{\textbf{W}}_{qk}} {\sqrt{d_k}})\textbf{1}) \\
& =
\mathrm{Diag}(\textbf{D}_0 \textbf{P} \textbf{1})) =  \mathrm{Diag}(\textbf{D}_0 \textbf{1})) = \textbf{D}_0 
\end{align} 
which indicates that \(\mathrm{exp}(\frac{\tilde{\textbf{W}}_{qk}}{\sqrt{d_k}}) = \textbf{D}_0 \textbf{P}  = \textbf{D}_{\tilde{\textbf{W}}_{qk}} \textbf{P}\). This completes the proof of \(d \geq n\) case of \textbf{Theorem 1}.

Regarding \(d < n\) case. Consider the special case of \(d=1\) and \(n=2\). There has \(\textbf{X} \in \mathcal{R}^{2 \times 1}\) and \(\textbf{W}_q, \textbf{W}_k \in \mathcal{R}^{1 \times 1}\). Suppose \(\textbf{X}=
\begin{bmatrix}
1 \\ 0 
\end{bmatrix}\), so that:
\begin{align}
\label{eqn: representation theorem proof 7}
\mathrm{Softmax}[\frac{(\textbf{X}\textbf{W}_q)(\textbf{X}\textbf{W}_k)^T}{\sqrt{d_k}}] & = \mathrm{Softmax}[\frac{\begin{bmatrix}
1 \\ 0 
\end{bmatrix}\textbf{W}_q\textbf{W}_k^T[1,0]}{\sqrt{d_k}}] \\
& = \mathrm{Softmax}[\begin{bmatrix}
\frac{\textbf{W}_q\textbf{W}_k^T}{\sqrt{d_k}} & 0\\ 
0 & 0
\end{bmatrix}]
\end{align} 

This matrix clearly cannot be used to generate \(\textbf{P}\). Then we extend the above special case to general values of \(n\) and \(d\), (\(d<n\)). Let \(\textbf{X}=[\textbf{1}_d,\cdots,\textbf{1}_d,\textbf{0}_d]^T=[\textbf{1}_{\mathrm{mat}},\textbf{0}_d]^T \in \mathcal{R}^{n \times d}\), where \(\textbf{1}_d,\textbf{0}_d \in \mathcal{R}^d\) denotes the all ones, zeros column vector, and \(\textbf{1}_{\mathrm{mat}}\) denotes the \(d \times (n -1)\) all ones matrix. Then
\begin{align}
\label{eqn: representation theorem proof 8}
\mathrm{Softmax}[\frac{(\textbf{X}\textbf{W}_q)(\textbf{X}\textbf{W}_k)^T}{\sqrt{d_k}}] & = \mathrm{Softmax}[\frac{[\textbf{1}_{\mathrm{mat}},\textbf{0}_d]^T \textbf{W}_q\textbf{W}_k^T[\textbf{1}_{\mathrm{mat}},\textbf{0}_d]}{\sqrt{d_k}}] \\
& = \mathrm{Softmax}[\begin{bmatrix}
\textbf{1}_{\mathrm{mat}}^T\frac{\textbf{W}_q\textbf{W}_k^T}{\sqrt{d_k}} \textbf{1}_{\mathrm{mat}} & \textbf{0}_{n-1}\\ 
\textbf{0}_{n-1} & 0
\end{bmatrix}]
\end{align} 

The basic idea of the proof remains consistency, and we can conclude the same conclusion.
According to the \textbf{Theorem 1}, we can not ensure the attention matrix of joint attention to be full rank.

\subsection{Proof of KMCTA can guarantee full rank}

\textbf{Assumption 1} The elements of \(\frac{(\textbf{X}\textbf{W}_q)(\textbf{X}\textbf{W}_k)^T}{\sqrt{d_k}}\) does not exist negative infinity in normal cases, which means that 0 will not appear at any other position in the attention matrix \(\tilde{\textbf{A}} = \mathrm{Softmax}[\frac{(\textbf{X}\textbf{W}_q)(\textbf{X}\textbf{W}_k)^T}{\sqrt{d_k}} + \tilde{ \textbf{M}}]\) except for the masked position. 

The softmax operation can ensure that the elements at all other positions are greater than 0 except for the masked position. The attention matrix \(\tilde{\textbf{A}}\) of KMCTA is always a lower triangular matrix due to the presence of the \(\tilde{ \textbf{M}}\) and the diagonal elements are always greater than 0, i.e., \(\forall a_{ii} > 0\):
\begin{equation}
\tilde{\textbf{A}} =
\begin{bmatrix}
a_{11} & 0 & 0 & \cdots & 0 \\
a_{21} & a_{22} & 0 & \cdots & 0 \\
a_{31} & a_{32} & a_{33} & \cdots & 0 \\
\vdots & \vdots & \vdots & \ddots & \vdots \\
a_{n1} & a_{n2} & a_{n3} & \cdots & a_{nn} \\
\end{bmatrix}
\end{equation}
The determinant of the \(\tilde{\textbf{A}}\), \(|\tilde{\textbf{A}}|= a_{11} a_{22} \cdots a_{nn} > 0 \neq 0\),  because \(\forall a_{ii} > 0\). Thus, the attention matrix \(\tilde{\textbf{A}}\) of KMCTA is always reversible, i.e. always be full rank.

\subsection{Proof of KMTA cannot guarantee full rank}

According to the \textbf{Assumption 1}. We consider a special case, when \(T\)=2 and \(L\)=2, the attention matrix \(\textbf{A} = \mathrm{Softmax}[\frac{(\textbf{X}\textbf{W}_q)(\textbf{X}\textbf{W}_k)^T}{\sqrt{d_k}} + \textbf{M}]\) of KMTA can be formulated as:
\begin{equation}
\begin{bmatrix}
a_{11} & 0 & a_{13} & a_{14} \\
0 & a_{22} & a_{23} & a_{24} \\
a_{31} & a_{32} & a_{33} & 0 \\
a_{41} & a_{42} & 0 & a_{44} \\
\end{bmatrix} \xrightarrow{} 
\begin{bmatrix}
a_{11} & 0 & a_{13} & a_{14} \\
0 & a_{22} & a_{23} & a_{24} \\
0 & 0 & a_{33} - \frac{a_{13}}{a_{11}} a_{31} - \frac{a_{23}}{a_{22}} a_{32} &  - \frac{a_{14}}{a_{11}} a_{31} - \frac{a_{24}}{a_{22}} a_{32} \\
0 & 0 & - \frac{a_{13}}{a_{11}} a_{41} - \frac{a_{23}}{a_{22}} a_{42} & a_{44}  - \frac{a_{14}}{a_{11}} a_{41} - \frac{a_{24}}{a_{22}} a_{42} \\
\end{bmatrix}
\end{equation}

where \(a_{11} + a_{13} + a_{14} = 1, a_{22} + a_{23} + a_{24} = 1, a_{31} + a_{32} + a_{33} = 1,a_{41} + a_{42} + a_{44} = 1\). When \(a_{33} - \frac{a_{13}}{a_{11}} a_{31} - \frac{a_{23}}{a_{22}} a_{32} = - \frac{a_{13}}{a_{11}} a_{41} - \frac{a_{23}}{a_{22}} a_{42}\), \(\textbf{A}\) can not always be full rank. For example:
\begin{equation}
\textbf{A} =
\begin{bmatrix}
0.4 & 0 & 0.4 & 0.2 \\
0 & 0.4 & 0.4 & 0.2 \\
0.4 & 0.5 & 0.1 & 0 \\
0.4 & 0.4 & 0 & 0.2 \\
\end{bmatrix}
\end{equation}
For a video, \(n=T \times L\), If we fix the number of attention heads, the dimension of tokens, the patch size, and the resolution of the input image, the only factor that can affect \(n\) for a video is the frame length. When we pray for a longer frame length and an increase in the number of tokens to improve performance, we may face the problems pointed out in \textbf{Theorem 1}. Because increasing the frame length will further increase \(n\), the expressive power of self attention may encounter bottlenecks. The low-rank bottleneck is vital, because it may lead to that many rows of the attention map are seriously homogenized. As the output of self-attention is the weighted sum of the same set of value vectors, the homogenization of attention weights inevitably leads to the resemblance among the aggregated features. 

\begin{figure*}[h!]
    \centering
    \includegraphics[width=0.8\textwidth,height=0.57\textwidth]{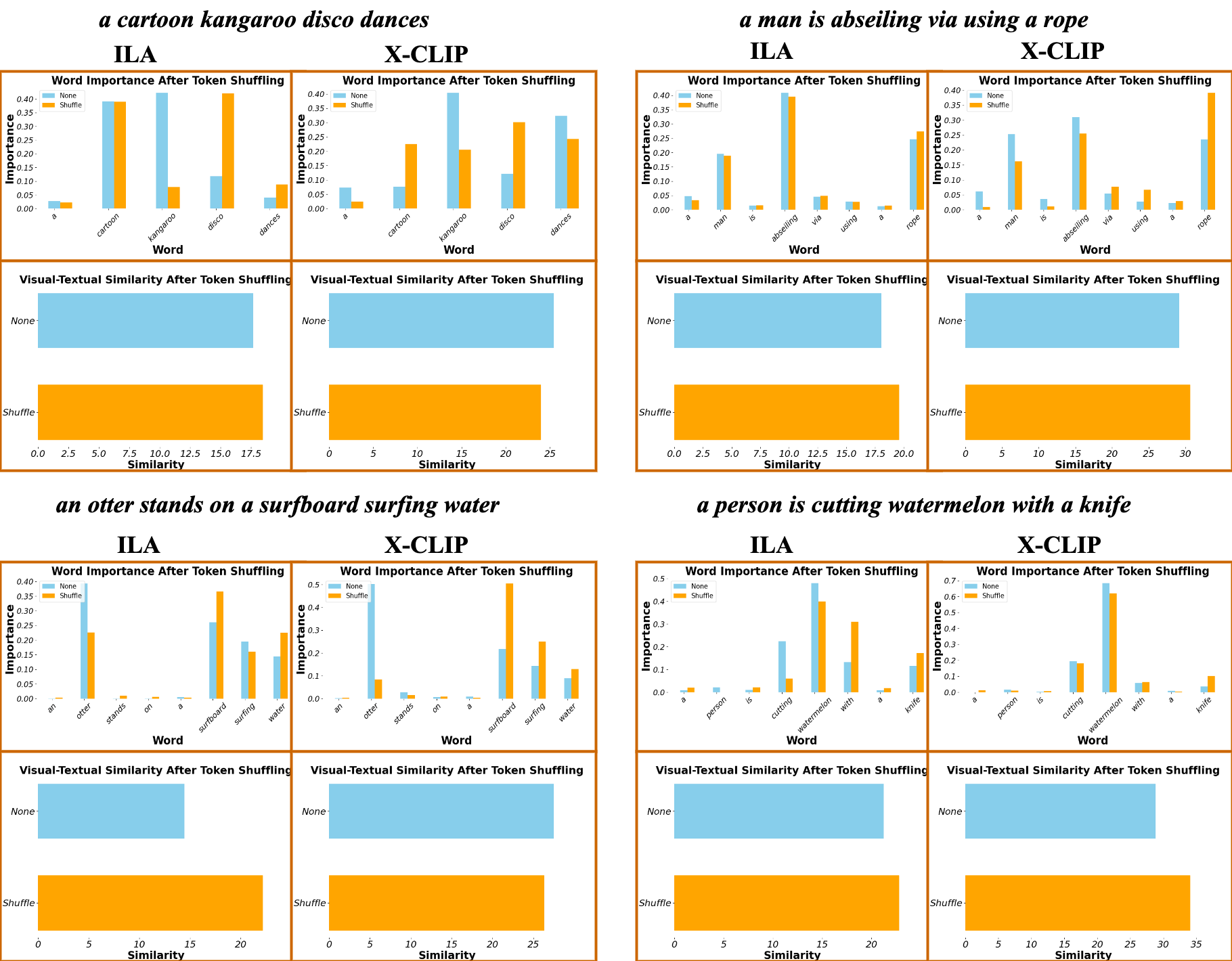}
    \caption{We visualize the impact of token shuffling on the ILA, XCLIP. In most cases, token shuffling has limited influence on their word importance. Surprisingly, in some figures, after token shuffling, the similarity actually increases, which is illogical. This indicates that the response of ILA and XCLIP to spatiotemporal position destruction is uncertain and has poor interpretability.}
    \label{fig: Spatiotemporal Homogenization ILA XCLIP}
\end{figure*}

\section{Spatiotemporal Homogenization Study}
\label{sec: Spatiotemporal Homogenization Study}
Fig. \ref{fig: Spatiotemporal Homogenizations} shows that equipping learnable position/time encoding with joint attention is not insufficient to alleviate spatiotemporal homogenization. For example, (Upper Left) 'a cartoon kangaroo disco dances'. We can observe that token shuffling has a limited impact on word importance and similarity of joint attention, but it has a significant impact on the word importance of KMT and both similarity and word importance of KMCT. (Upper Right) 'a man is abseiling via using a rope'. (Lower Left) 'an otter stands on a surfboard surfing water'. (Lower Right) 'a person is cutting watermelon with a knife'. The key point is that classification accuracy does not necessarily indicate that these learnable position encoding or attention biases genuinely grasp the concept of spatiotemporal structure. Since the optimization goal of the network is usually to pursue performance, the learned positional encoding or attention bias serve to improve accuracy rather than understand the spatiotemporal concept. This is because their optimization is guided by gradients aimed at increasing accuracy, rather than by an objective function that tells them what spatiotemporal structure is. The Kronecker mask in KMTA and KMCTA pursues not only for good performance, but also for interpretability. The Kronecker mask act as a natural spatiotemporal heterogeneity inductive bias, KMTA/KMCTA exhibit better spatiotemporal complementarity with spatial attention. Additionally, in Fig. \ref{fig: Spatiotemporal Homogenization ILA XCLIP}, we show that several tailored modules (ILA and XCLIP) also encounter spatiotemporal homogenization issues during our testing. Surprisingly, in some cases like 'a man is abseiling via using a rope,' 'an otter stands on a surfboard surfing water,' and 'a person is cutting watermelon with a knife,' the similarity actually increases after token shuffling, which is illogical.

Additionally, we evaluate the performance of joint attention, Kronecker mask temporal attention (KMTA) and Kronecker mask causal temporal attention (KMCTA) on K400 \cite{kay2017kinetics} after introducing token shuffling. The results are shown in Tab. \ref{table: token shuffling}. The greater performance degradation observed in KMTA and KMCTA indicates that they are more sensitive to disturbances in the spatiotemporal structure, thus alleviating spatiotemporal homogenization.

\begin{table}[h!]
\vspace{-0.2cm}
        \centering
        \caption{Performance of joint attention, KMTA and KMCTA while introducing token shuffling.}
\vspace{-0.2cm}
        \label{table: token shuffling}
 \resizebox{0.48\textwidth}{!}{   
        \begin{tabular}{cc}
            \toprule
\textbf{Token Shuffling (Yes-Y, No-N)} & \textbf{Top-1 (\%)} \\
\hline
Joint Attention, N & 80.1\\
Kronecker Mask Temporal Attention, N & 81.5\\
Kronecker Mask Causal Temporal Attention, N & 81.4\\
Joint Attention, Y & 58.9 (-21.2)\\
Kronecker Mask Temporal Attention, Y & 43.5 (-38.0)\\
Kronecker Mask Causal Temporal Attention, Y & 42.6 (-38.8)\\
\bottomrule
        \end{tabular}
}
\vspace{-0.3cm}
\end{table}


\section{Additional Experiments}
\label{sec: Supplementary Experiments}
\textbf{The effects of interpretive prompt.} In Tab. \ref{table: comparison text prompts}, we can observe that, directly use verbs and phrases achieve the lowest performance. When employing prefix and suffix prompt templates, the performance slightly improved. When introducing action decomposition interpretive prompts, the performance significantly improved. When adding synonym conversion interpretive prompts, the performance is further increased. Last, we involve body parts interpretive prompts that the performance reaches the best. 

\textbf{Which branches should be finetuned?} We separately freeze the parameters of the pretrained image and text encoder. From Tab. \ref{table: finetune}, we conclude the following observations: 1) Freezing both image and text encoders, only tuning KMT transformer achieves the worst performance. 2). Finetuning only one of them will improve the performance. 3). Finetuning them simultaneously achieves the best performance.

\textbf{The Evaluation performance of zero- and few-shot experiments on Kinetics-600, HMDB-51, and UCF-101 with CLAVER-B/16 (KMCT) configuration.} The comparison of KMT and KMCT as shown in Tab. \ref{table: comparison KMT and KMCT zero-shot}, \ref{table: comparison KMT and KMCT few-shot}. In the zero-shot scenario, the performance of KMCTA and KMTA has its own wins and losses, which is due to the lack of further fine-tuning, and the model's ability is relatively dependent on previous training. In the few-shot scenario, we can observe that in most cases, the performance of KMCTA is better, which further indicates that when the scale of training data is limited, the low-rank bottleneck is more significant.
\begin{table}[h!]
\vspace{-0.2cm}
    \begin{minipage}[t!]{0.5\textwidth}
        \centering
        \caption{Ablation on interpretive prompt.}
        \vspace{-0.2cm}
        \label{table: comparison text prompts}
 \resizebox{0.95\textwidth}{!}{   
        \begin{tabular}{ccc}
            \toprule
\textbf{Text Prompts}& \textbf{Top-1 (\%)} & \textbf{Top-5 (\%)} \\
\hline
Noun and Phrase of Action & 77.4& 93.5\\
$+$ Prefix-Suffix & 77.4& 93.6\\
$+$ Action Decomposition Interpretive Prompt& 78.0& 94.0\\
$+$ Synonym Conversion Interpretive Prompt& 78.2& 94.2\\
$+$ Involving Body Parts Interpretive Prompt& \textbf{78.4}& \textbf{94.3}\\
\bottomrule
        \end{tabular}
}
    \end{minipage}
  \begin{minipage}[t!]{0.5\textwidth}
        \centering
        \caption{Which branches should be finetuned.}
        \vspace{-0.2cm}
       \label{table: finetune}
            \resizebox{0.98\textwidth}{!}{   
        \begin{tabular}{ccccc}
            \toprule
\textbf{KMT transformer}& \textbf{Visual}& \textbf{Text}& \textbf{Top-1 (\%)}& \textbf{Top-5 (\%)}\\
\hline
$\checkmark$& $\times$& $\times$& 78.7& 94.4\\
$\checkmark$& $\checkmark$& $\times$& 81.0& 95.1\\
$\checkmark$& $\times$& $\checkmark$& 80.8& 95.2\\
$\checkmark$& $\checkmark$& $\checkmark$& \textbf{81.5}& \textbf{95.5}\\
\bottomrule
        \end{tabular}
}
    \end{minipage}
\vspace{-0.3cm}
\end{table}

\begin{table}[h!]
\vspace{-0.2cm}
    \begin{minipage}[t!]{0.5\textwidth}
        \centering
        \caption{Comparison of KMT and KMCT on zero-shot settings.}
        \vspace{-0.2cm}
        \label{table: comparison KMT and KMCT zero-shot}
 \resizebox{0.95\textwidth}{!}{   
        \begin{tabular}{cccc}
            \toprule
\textbf{Zero-shot} & \textbf{Kinetics-600 (\%)} & \textbf{HMDB-51 (\%)} & \textbf{UCF-101 (\%)} \\
\hline
CLAVER-B/16 (KMT) & 73.8 $\pm$ 0.6 & \textbf{54.1} $\pm$ 2.4& \textbf{78.6} $\pm$ 1.7\\
CLAVER-B/16 (KMCT) & \textbf{74.1} $\pm$ 0.9& 54.0 $\pm$ 2.0 & 78.4 $\pm$ 2.1 \\
\bottomrule
        \end{tabular}
}
    \end{minipage}
  \begin{minipage}[t!]{0.5\textwidth}
        \centering
        \caption{Comparison of KMT and KMCT on few-shot settings.}
        \vspace{-0.2cm}
       \label{table: comparison KMT and KMCT few-shot}
            \resizebox{0.98\textwidth}{!}{   
        \begin{tabular}{ccccccccc}
            \toprule
\textbf{CLAVER-B/16} & \multicolumn{4}{c}{\textbf{KMT}} & \multicolumn{4}{c}{\textbf{KMCT}} \\
\hline
Few-shot & K=2 & K=4 & K=8 & K=16 & K=2 & K=4 & K=8 & K=16\\
HMDB-51 (\%) & \textbf{58.6} & 63.9 & 68.0 & 72.5 & 58.3 & \textbf{64.5} & \textbf{68.6} & \textbf{72.9} \\
UCF-101 (\%) & 89.7 & \textbf{92.9} & 96.1 & 98.0 & \textbf{90.0} & 92.9 & \textbf{96.6} & \textbf{98.1}\\
\bottomrule
        \end{tabular}
}
    \end{minipage}
\vspace{-0.3cm}
\end{table}

\section{Experimental Setting Details}
\label{sec: Experiment Setting Details}
\textbf{Architectures.} CLAVER-B/32 adopts ViT-B/32 (\(L_V\)=12, \(N_{h}\)=12, \(d\)=768, \(p\)=32) and is equipped with a KMT/KMCT transformer (\(L_K\)=2, \(N_{h}\)=12, \(d\)=768, \(p\)=32). CLAVER-B/16 employs ViT-B/16 (\(L_V\)=12, \(N_{h}\)=12, \(d\)=768, \(p\)=16), along with the KMT/KMCT transformer (\(L_K\)=2, \(N_{h}\)=12, \(d\)=768, \(p\)=32). CLAVER-L/14 is equipped with ViT-L/14 (\(L_V\)=24, \(N_{h}\)=16, \(d\)=1024, \(p\)=14) and the KMT/KMCT transformer (\(L_K\)=4, \(N_{h}\)=16, \(d\)=1024, \(p\)=14). Here \(L_V\) denotes the layers of ViT, \(L_K\) denotes the layers of Kronecker mask temporal transformer, \(N_h\) refers to the number of attention heads, \(d\) represents the embedding dimension and \(p\) is the patch size.

\noindent
\textbf{Hyperparameters.} The experiments are conducted on 8 NVIDIA 80G A100 GPUs. We present the training hyperparameters in Tab. \ref{table: hyperparameters setting}. Additionally, the learning rate for updating the KMT/KMCT transformer (randomly initialized) parameters is set 10$\times$ higher than the learning rate for parameters of the text encoder or image encoder. Because the text/image encoder already possesses a ability to extract high-quality text/image representations, conversely, KMT/KMCT transformer is trained from scratch. In the experiment, we freeze the parameters of the CLIP's image encoder to reduce certain computational costs, as it already has strong image feature extraction capabilities. And then we also conducted experiments on this aspect in subsequent ablation study.

\begin{table*}[h!]
\caption{The training hyperparameters settings of experiments.}
\label{table: hyperparameters setting}
\centering
\resizebox{0.8\textwidth}{!}{
\begin{tabular}{b{3.5cm}b{4cm}b{4cm}b{4cm}}
\hline
\multicolumn{1}{c}{\textbf{Config}} & \multicolumn{1}{c}{\textbf{Fully-sup}} & \multicolumn{1}{c}{\textbf{Few-shot}} & \multicolumn{1}{c}{\textbf{Zero-shot}} \\
\hline
\multicolumn{1}{c}{Optimizer} & \multicolumn{3}{c}{AdamW} \\ 

\multicolumn{1}{c}{Base learning rate} & \multicolumn{1}{c}{12e-6} & \multicolumn{1}{c}{2e-6} & \multicolumn{1}{c}{12e-6} \\

\multicolumn{1}{c}{Minimal learning rate} & \multicolumn{1}{c}{12e-8} & \multicolumn{1}{c}{2e-8} & \multicolumn{1}{c}{12e-8} \\

\multicolumn{1}{c}{Weight decay} & \multicolumn{3}{c}{0.001} \\

\multicolumn{1}{c}{Optimizer betas} & \multicolumn{3}{c}{$\beta_1$, $\beta_2$ =0.9, 0.98}\\

\multicolumn{1}{c}{Batch size} & \multicolumn{3}{c}{128 (ViT-B) 32 (ViT-L)} \\

\multicolumn{1}{c}{Learning rate schedule} & \multicolumn{3}{c}{Cosine decay} \\

\multicolumn{1}{c}{Warmup epochs} & \multicolumn{3}{c}{5} \\

\multicolumn{1}{c}{Training epochs} & \multicolumn{1}{c}{60 (ViT-B) 40 (ViT-L)} & \multicolumn{1}{c}{80 (20 on K400)} & \multicolumn{1}{c}{0 (20 on K400)} \\

\multicolumn{1}{c}{\multirow{2}{*}{Augmentation}} & \multicolumn{3}{c}{RandomFlip, MultiScaleCrop, ColorJitter} \\
 & \multicolumn{3}{c}{GrayScale, Label smoothing, Mixup, Cutmix} \\
 \hline
\end{tabular}
}
\end{table*}


\textbf{Fully-supervised experiments setting.} We conduct the fully-supervised experiments on Kinetics-400\&600. During training, a sparse sampling strategy is used. The number of frames is set to 8 or 16. We spatially scale the shorter side of each frame to 256 and take a 224 crop center crop. Following, we adopt the multi-view inference with 3 spatial crops and 4 temporal clips.

\textbf{Few-shot experiments setting.} We randomly sample 2, 4, 8 and 16 videos from each class on UCF-101 and HMDB-51 constructing the training set. For evaluation, we use the first split of the test set on UCF-101 and HMDB-51.

\textbf{Zero-shot experiments setting.} We train CLAVER-B/16 with 32 frames on Kinetics-400. The same as, we apply the following two evaluation protocols in zero-shot experiments. 1) Evaluation for HMDB-51 and UCF-101.  Following, the prediction is conducted on the three splits of the test data, and we report the average top-1 accuracy and standard deviation. 2) Evaluation for Kinetics-600. Following, the 220 new categories outside Kinetics-400 in Kinetics-600 are used for evaluation. The evaluation is conducted three times. For each iteration, we randomly sampled 160 categories for evaluation from the 220 categories in Kinetics-600.

\section{Interpretive Prompt Details}
\label{sec: Interpretive Prompt Details}
First, we input command prompts into the ChatGPT \cite{openai2024gpt4technicalreport} chatbot API to obtain a demonstration case of a small amount (in our experiment, we set it to 3 here) of detailed descriptions of actions.

\noindent
\textbf{Question to ChatGPT:} 
\begin{lstlisting}
Below, I will provide you with some action nouns. Please provide a simple 
and detailed description (explanation) about the action decomposition
of these action nouns. Please note that the sentence length of the 
description should not exceed 76 words. 
\end{lstlisting}

\noindent
\textbf{Answered by ChatGPT:}
\begin{lstlisting}
Certainly! Please go ahead and provide the action nouns you'd like 
descriptions for, and I'll provide concise explanations within the 
specified sentence length.
\end{lstlisting}

\noindent
\textbf{Question to ChatGPT:}
\begin{lstlisting}
abseiling.
\end{lstlisting}

\noindent
\textbf{Answered by ChatGPT:}
\begin{lstlisting}
Abseiling combines several actions to descend a vertical surface with a 
rope. Climbers secure themselves with a harness and utilize a descender 
device for controlled descent. Simple actions, like maintaining a 
straight body position and regulating rope tension, form the basis. 
Abseiling demands proper training, safety measures, and is popular in 
adventure sports and rescue operations, allowing individuals to 
experience controlled descent in various settings.
\end{lstlisting}

\noindent
\textbf{Question to ChatGPT:}
\begin{lstlisting}
air drumming.
\end{lstlisting}

\noindent
\textbf{Answered by ChatGPT:}
\begin{lstlisting}
Air drumming is a rhythmic expression where individuals simulate playing 
drums without physical instruments. Simple actions, like mimicking 
drumming motions in the air, combine to create this imaginative and 
playful activity. Enthusiasts use their hands and feet to imitate 
drumming patterns, syncing with music. It's a spontaneous, enjoyable 
gesture often done during music listening or live performances, 
showcasing one's connection to the rhythm without the need for actual 
drums or drumsticks.
\end{lstlisting}

\noindent
\textbf{Question to ChatGPT:}
\begin{lstlisting}
answering questions.
\end{lstlisting}

\noindent
\textbf{Answered by ChatGPT:}
\begin{lstlisting}
Answering questions involves providing responses to queries posed by 
others. Simple actions like active listening, comprehension, and concise 
articulation combine in this communicative process. It is fundamental 
in various contexts, facilitating information exchange and problem-
solving. Respondents draw on their knowledge and expertise to address 
inquiries, contributing to effective communication and fostering 
understanding between individuals or groups.
\end{lstlisting}

\noindent
\textbf{Question to ChatGPT:}
\begin{lstlisting}
Below, I will provide you with some action nouns or phrases. Please 
provide many words and phrases that share the same central concept 
as these action nouns or phrases but have more diverse expressions. 
Please note that your reply should not exceed 76 words in length.
\end{lstlisting}

\noindent
\textbf{Answered by ChatGPT:}
\begin{lstlisting}
1. Running: Jogging, sprinting, dashing, hurrying, racing, pounding the pavement, speeding, trotting.
2. Eating: Dining, feasting, devouring, munching, nibbling, gobbling up, indulging, savoring.
3. Writing: Scribbling, jotting down, composing, drafting, penning, inscribing, scrawling, scripting.
4. Laughing: Chuckling, giggling, guffawing, snickering, roaring with laughter, belly laughing, cackling, chortling.
5. Reading: perusing, browsing, scanning, skimming, devouring books, absorbing, immersing oneself, delving into texts.
\end{lstlisting}

\noindent
\textbf{Question to ChatGPT:}
\begin{lstlisting}
Cutting in the kitchen.
\end{lstlisting}

\noindent
\textbf{Answered by ChatGPT:}
\begin{lstlisting}
Slicing, dicing, chopping, mincing, cleaving, carving, trimming, 
preparing ingredients.
\end{lstlisting}

\noindent
\textbf{Question to ChatGPT:}
\begin{lstlisting}
driving car.
\end{lstlisting}

\noindent
\textbf{Answered by ChatGPT:}
\begin{lstlisting}
Operating a vehicle, maneuvering behind the wheel, navigating the road, 
piloting an automobile, steering, cruising, commuting by car, motoring.
\end{lstlisting}

\noindent
\textbf{Question to ChatGPT:}
\begin{lstlisting}
Walking With Dog.
\end{lstlisting}

\noindent
\textbf{Answered by ChatGPT:}
\begin{lstlisting}
Strolling with a canine companion, ambling with a pet, promenading with a 
pup, hiking with a furry friend, sauntering alongside a dog, wandering 
with a four-legged buddy, leash-walking, trotting with a pooch.
\end{lstlisting}

\noindent
\textbf{Question to ChatGPT:}
\begin{lstlisting}
Below, I will provide you with some action nouns. Please describe 
these actions based on their nouns and possible body parts involved. 
Please note that the sentence length of your response should not exceed 
76 words.
\end{lstlisting}

\noindent
\textbf{Answered by ChatGPT:}
\begin{lstlisting}
1. Running: The rhythmic motion of legs propelling the body forward, feet striking the ground with each stride, muscles contracting and relaxing to generate momentum.
2. Swimming: Arms slicing through water, legs kicking to maintain propulsion, torso rotating to streamline movement, lungs rhythmically inhaling and exhaling for oxygen.
3. Writing: Fingers gripping a pen or pencil, hand moving across paper or keyboard, muscles in the forearm and wrist coordinating precise movements to form letters and words.
4. Jumping: Legs bending at the knees, feet pushing off the ground, arms swinging for momentum, body soaring through the air momentarily before landing.
5. Dancing: Feet tapping, legs extending and bending, hips swaying, arms gesturing and flowing in sync with music, body moving rhythmically with expression and grace.
\end{lstlisting}

\noindent
\textbf{Question to ChatGPT:}
\begin{lstlisting}
Cutting in the kitchen.
\end{lstlisting}

\noindent
\textbf{Answered by ChatGPT:}
\begin{lstlisting}
Using a sharp knife, fingers gripping the handle, hand guiding the blade 
through ingredients on a cutting board, wrist controlling the motion, 
fingers curling slightly to hold the food steady, precision applied 
to achieve desired shapes or sizes, ensuring safety and efficiency 
during food preparation.
\end{lstlisting}

\noindent
\textbf{Question to ChatGPT:} 
\begin{lstlisting}
driving car.
\end{lstlisting}

\noindent
\textbf{Answered by ChatGPT:} 
\begin{lstlisting}
Gripping the steering wheel, hands adjusting position, fingers pressing 
pedals for acceleration and braking, eyes scanning surroundings for 
obstacles, feet coordinating between clutch, brake, and accelerator, 
body positioned comfortably in the driver's seat, mind focused on navigation 
and traffic signals, reacting swiftly to changing road conditions.
\end{lstlisting}

\noindent
\textbf{Question to ChatGPT:}
\begin{lstlisting}
Walking With Dog.
\end{lstlisting}

\noindent
\textbf{Answered by ChatGPT:} 
\begin{lstlisting}
Leash in hand, fingers securing grip, arm relaxed as it swings alongside 
the body, legs moving in tandem with the dog's pace, feet stepping 
forward with purpose, eyes attentive to the dog's behavior and surroundings, 
occasional stops for sniffing or marking, a bond of companionship 
evident in synchronized movement.
\end{lstlisting}

Subsequently, as illustrated in Fig \ref{fig: Interpretive Prompt}, we input the examples generated by ChatGPT \cite{openai2024gpt4technicalreport} into LLaMA-3 \cite{dubey2024llama3herdmodels} multiple times in the format of \textit{\textbf{Command}} \(+\) \textit{\textbf{Examples}} \(+\) \textit{\textbf{Action Concept}}, automatically generating diverse text descriptions about the \textit{\textbf{Action Concept}} through the program. When dealing with datasets containing a large number of categories, this scheme can greatly save labor costs. In addition, it should be noted that LLaMA-3 offers controllable parameters to control the randomness (\(\tau\)) and diversity (\(p\)) of output texts. We set \(\tau\) and \(p\) to 0.90 and 0.95, respectively to ensure that there are significant variations in output content for the same input, while maintaining consistency in the central concept.


\section{Analysis of Synthetic Video Testing}
\label{sec:Analysis of Synthetic Video Testing}
\begin{figure*}[h!]
    \centering
    \includegraphics[width=0.85\textwidth,height=0.36\textwidth]{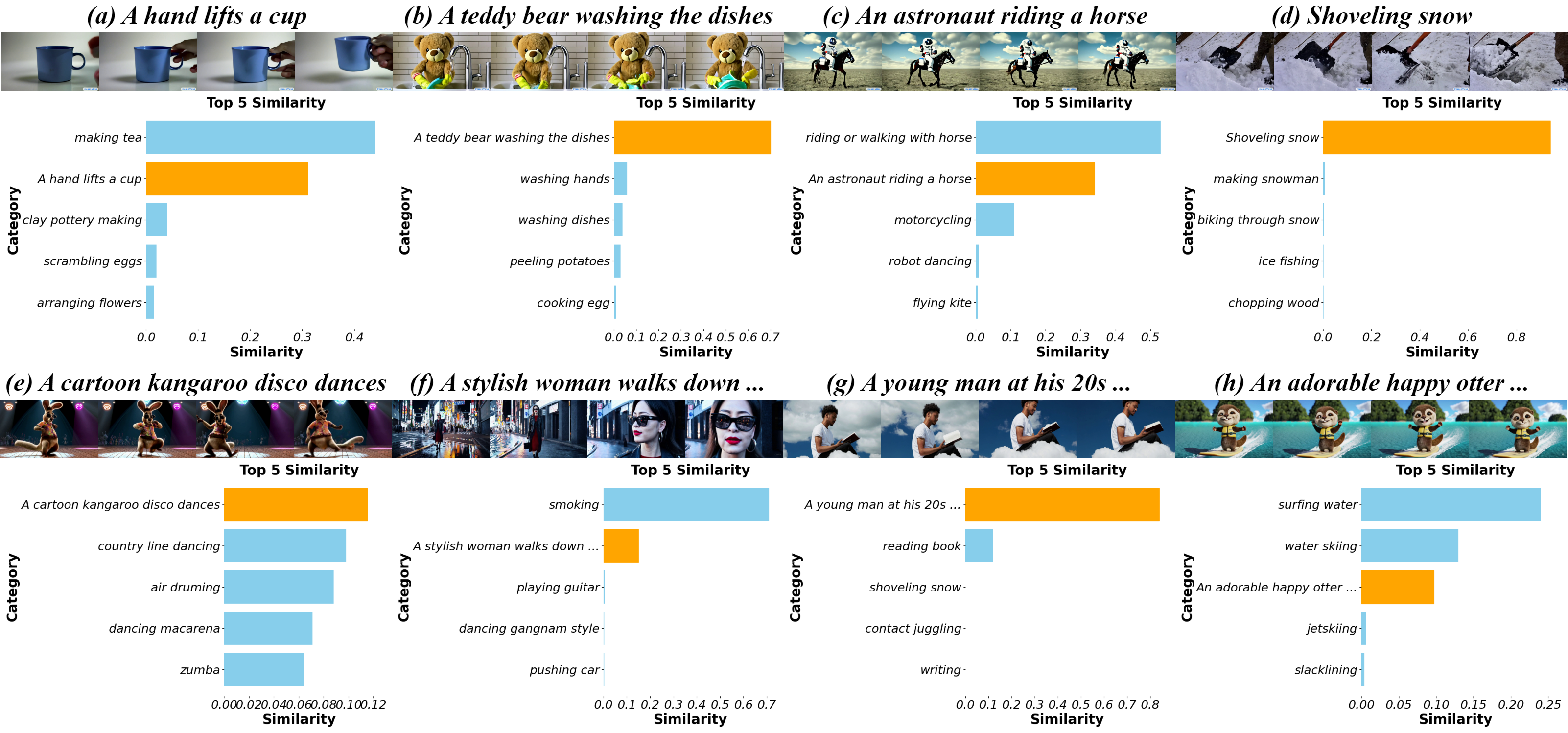}
    \vspace{-0.3cm}
    \caption{Testing on synthetic videos from Imagen (\textbf{Upper}) and Sora (\textbf{Lower}). The bar chart represents the Top-5 similarity texts, orange bar denotes the original prompts of synthetic videos.}
    \label{fig: Generative model Imagen Sora  generate samples}
\vspace{-0.7cm}
\end{figure*}

We use the original prompts of synthetic videos (several action-related examples generated by Imagen \cite{saharia2022photorealistic} and Sora \cite{videoworldsimulators2024}) as their corresponding text descriptions, and show Top-5 most relevant texts in Fig. \ref{fig: Generative model Imagen Sora  generate samples}. The results shows the robustness and generalization of CLAVER.

For cases generated by Imagen, in Fig. \ref{fig: Generative model Imagen Sora  generate samples} \textbf{Upper}, we denote the Kinetics-400 action text set as \(\mathcal{A}\). The action category "lift a cup" in Fig. \ref{fig: Generative model Imagen Sora  generate samples} (a) does not appear in \(\mathcal{A}\), while, (b) "washing dishes", (c) "riding horse", (d) ''shoveling snow" do appear in \(\mathcal{A}\). All videos do not come from the real-world, which are synthetic. We observe that all the original prompts appear in the Top-5 similarity (more precisely, Top-2). Specifically, for (a), "making tea" has the highest similarity, possibly due to the presence of cups and human hands in the scene. In addition, the original text has the second-highest similarity. In (b), apart from the original prompt in Top-1, other texts such as "washing dishes" convey the same meaning as the original prompts but lack a description of the subject ("teddy") of the action. "Washing hands" and "peeling potatoes" may involve actions and scenes like hand movements and sinks, thus exhibiting high similarity. For (c), although the one with the highest similarity is not "An astronaut riding a horse" , "riding or walking horses" is the exactly describes the action in the video, with the only difference being the lack of the subject of the action. For (d), "Shoveling snow" belongs to \(\mathcal{A}\), and since the original prompt of the video is "Shoveling snow", its similarity is very high, exceeding 90\%. 

For the cases generated by Sora, as depicted in Fig. \ref{fig: Generative model Imagen Sora  generate samples} \textbf{Lower}, (e) "disco dances" concept is not in \(\mathcal{A}\) but has the highest similarity. Besides, other action concepts in the Top-5 ("country line dancing","air druming", "dancing macarena","zumba") are all dance-related. (f) is a relatively challenging sample here. Although the ground truth description is located in the second, the "smoking" in Top-1 is entirely unrelated to the content of the video, because the original prompt of the video primarily describes the scene and the action-related text is relatively short. In addition, the shot is a process from far to near, which may lead the model to arrive at an unreasonable Top-1. In (g), the Top-2 descriptions are original prompt and "reading book", both strongly related to the action in video. For (h), the original prompt only rank third, however, Top-1 "surface water" and the second "water skimming" are descriptions strongly related to the action in video.

\section{Dataset Details}
\label{sec: Dataset Details}
\textbf{\textit{Kinetics-400\&600}.} The Kinetics dataset consists of 10-second video clips collected from YouTube. Specifically, Kinetics-400 \cite{kay2017kinetics} consists of approximately 240k training videos and 20k validation videos with 400 categories, while Kinetics-600 \cite{carreira2018short} is an extension of Kinetics-400, consisting of approximately 410k training videos and 29k validation videos with 600 categories.

\noindent
\textbf{\textit{UCF-101}.} UCF-101 \cite{soomro2012ucf101} consists of 101 action categories, over 13k clips and 27 hours of video data. The database comprises realistic user uploaded videos containing camera motion and cluttered backgrounds. The training and test data are divided into three splits.

\noindent
\textbf{\textit{HMDB-51}.} HMDB-51 \cite{kuehne2011hmdb} is a collection of realistic videos from various sources, including movies and web videos. It is composed of 6,766 video clips from 51 action categories, with each category containing at least 101 clips. The dataset is divided into three splits for training and test data. In each split, each action class has 70 clips for training and 30 clips for testing. 


\noindent
\textbf{\textit{Synthetic Videos}.} We select some action-related videos from the demo cases generated by Imagen \cite{saharia2022photorealistic} and Sora \cite{videoworldsimulators2024}, and employ the original prompts and their corresponding synthetic videos as a pair of test samples.

\section{More Attention Heat Map Visualization}
\label{sec: More Attention Heat Map Visualization}
We visualize several spatiotemporal attention map of samples from Kinetics400, Imagen, Sora.

\begin{figure*}[h!]
    \centering
    \includegraphics[width=0.79\textwidth,height=1.4\textwidth]{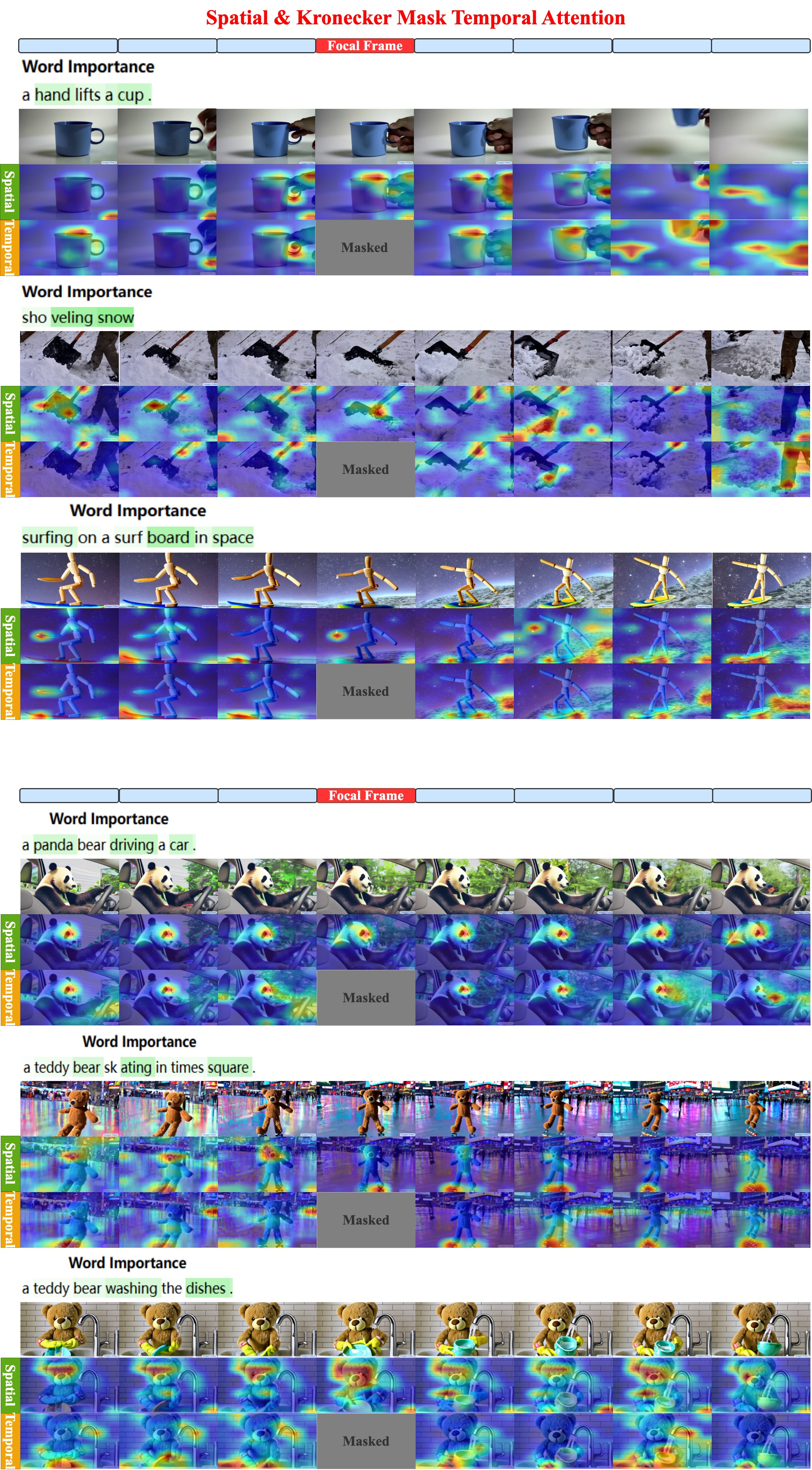}
    \caption{Attention map of synthetic videos from Imagen.}
    \label{fig: Imagen Attention Map}
\end{figure*}

\begin{figure*}[h!]
    \centering
    \includegraphics[width=0.79\textwidth,height=0.71\textwidth]{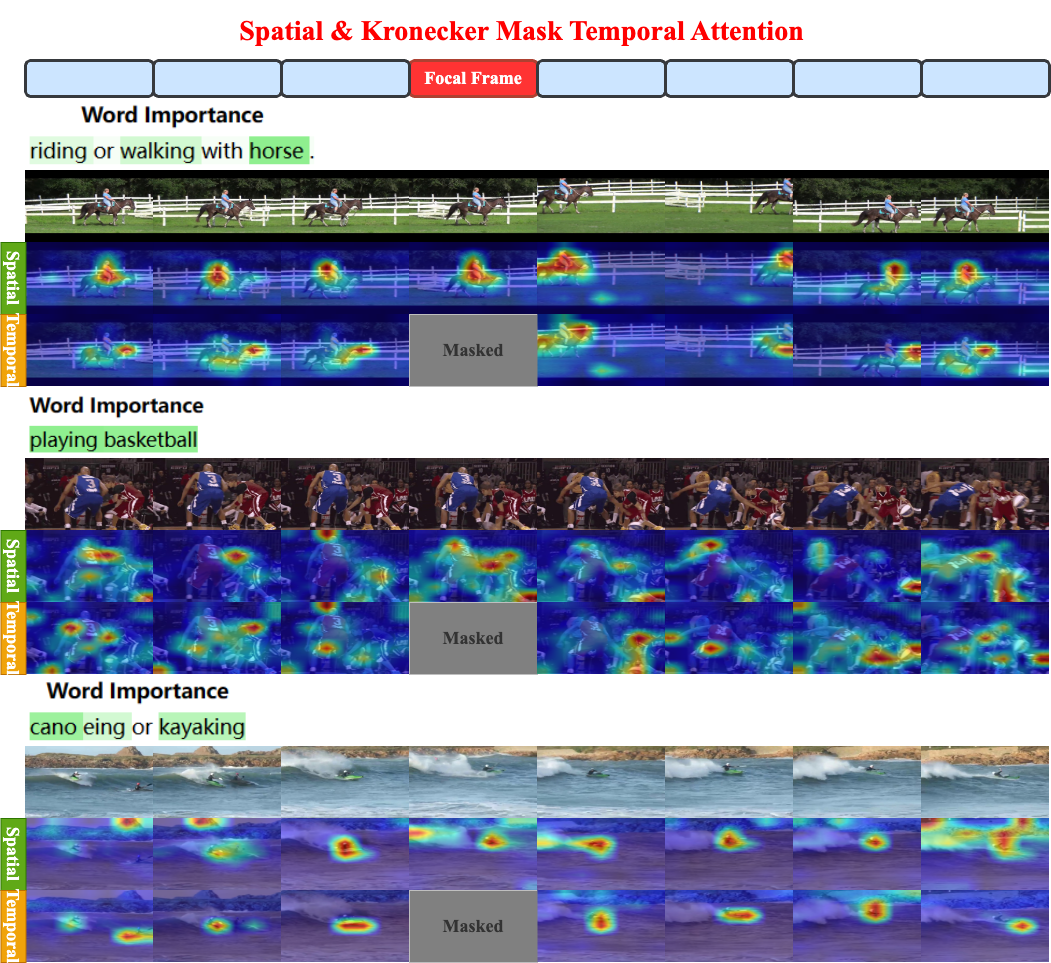}
    \caption{Attention map of videos from Kinetics400.}
    \label{fig: Samples Attention Map}
\end{figure*}

\begin{figure*}[h!]
    \centering
    \includegraphics[width=0.79\textwidth,height=0.71\textwidth]{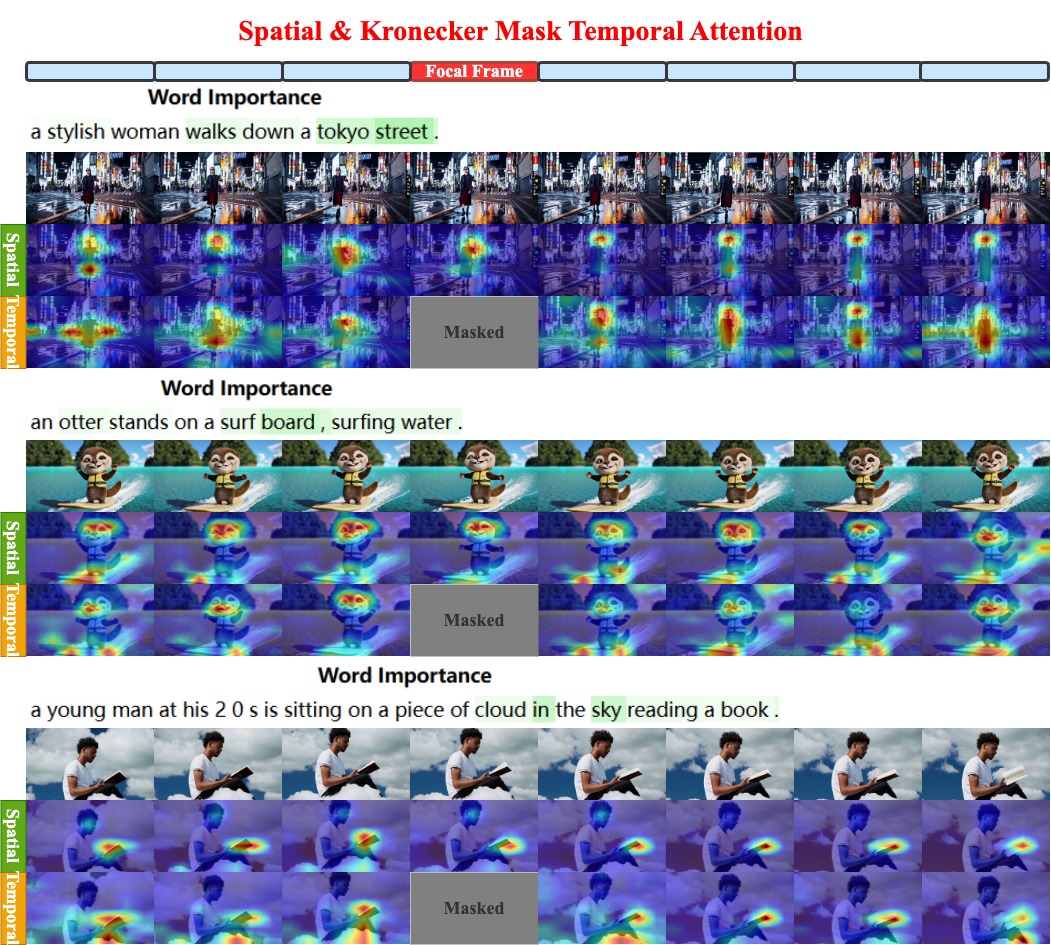}
    \caption{Attention map of synthetic videos from Sora.}
    \label{fig: Sora Attention Map}
\end{figure*}

\begin{figure*}[t]
\centering
\includegraphics[width=0.89\textwidth,height=0.27\textwidth]{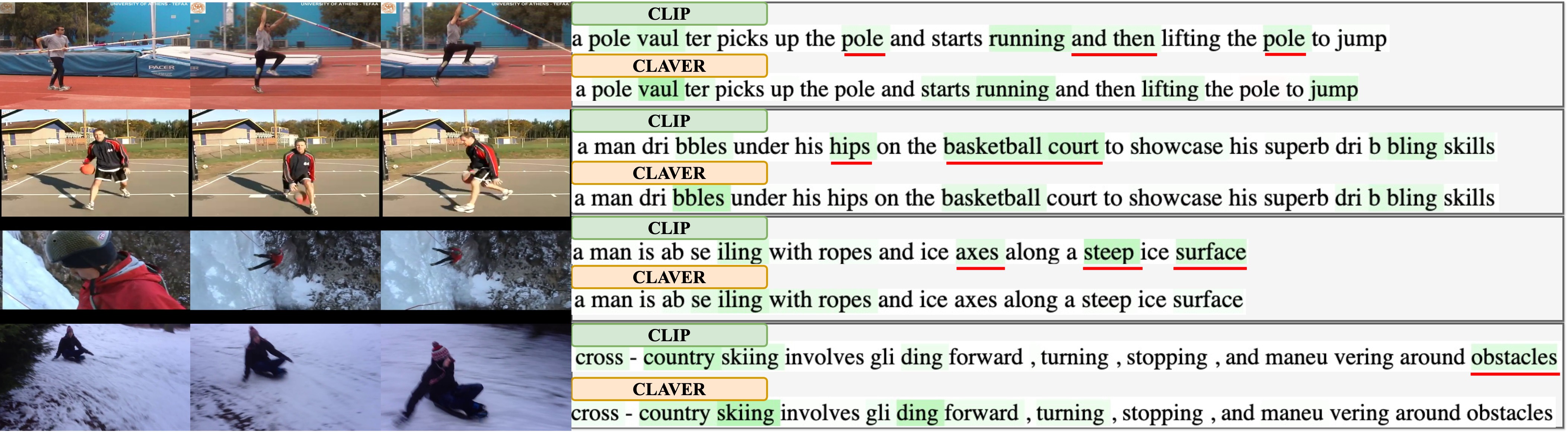}
\caption{More visualization examples of transition of CLIP's attention on nouns to CLAVER's preference on verbs.}
    \label{fig: Nouns to verbs}
\end{figure*}

\begin{figure*}[h!]
    \centering
    \includegraphics[width=0.87\textwidth,height=0.9\textwidth]{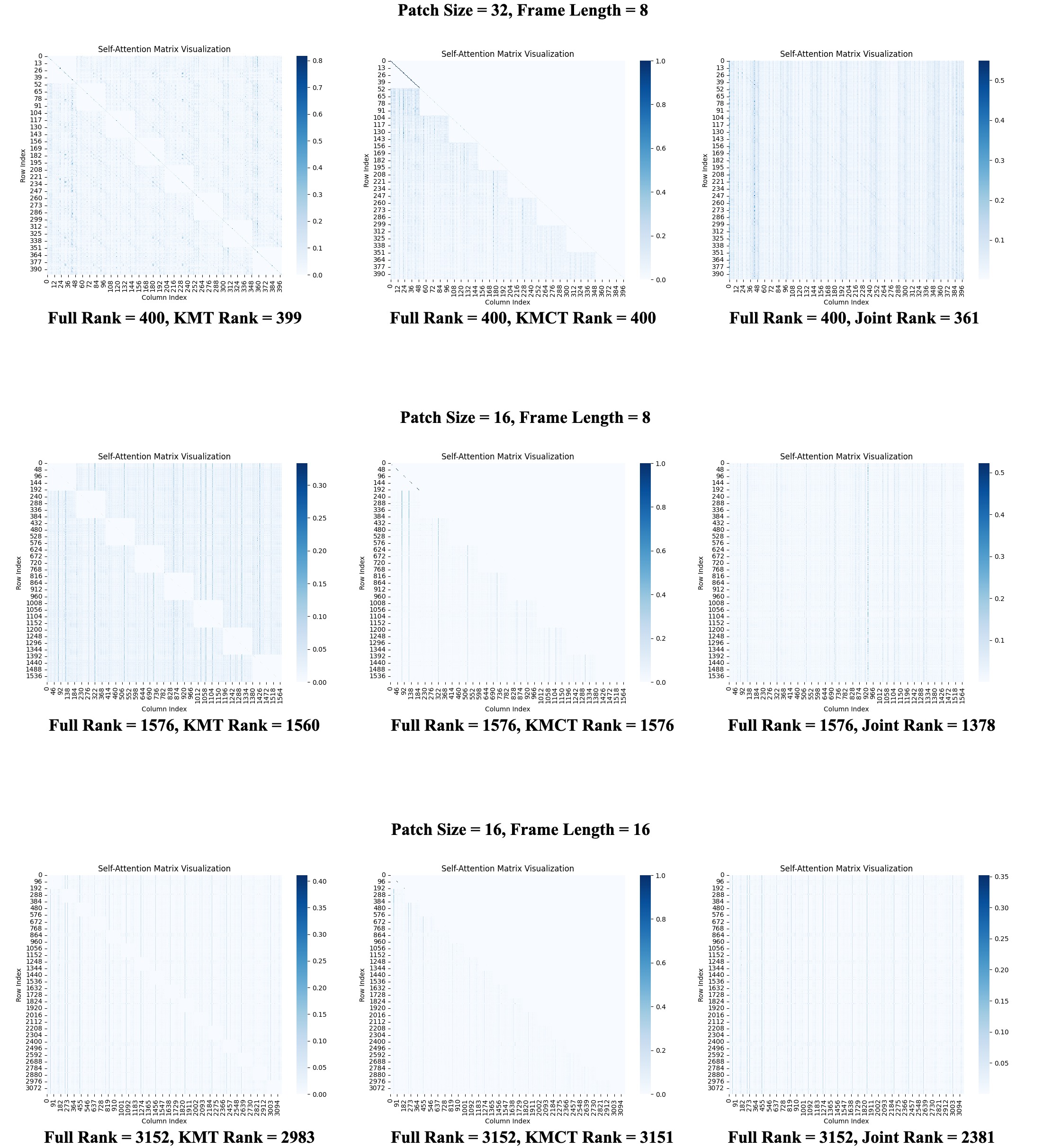}
    \caption{The rank of Kronecker temporal attention (KMT), Kronecker causal temporal attention (KMCT) and joint attention (Joint) on different patch size and frame length. (The calculation of rank is done via torch.linalg.svd)}
    \label{fig: Attention Matrix}
\end{figure*}


\end{document}